\documentclass[11pt,openany]{book}

\title{Mémoire Sihem}

\usepackage{pagenote}
\usepackage{graphicx}
\usepackage[utf8]{inputenc}
\usepackage{tabularx,ragged2e,booktabs}
\usepackage{amsmath,amsthm,amssymb,latexsym}
\usepackage{mathrsfs}
\usepackage{setspace}
\usepackage{enumitem}
\usepackage{titlesec}
\usepackage[LFE,LAE, T1]{fontenc}
\usepackage{algorithm}
\usepackage{algpseudocode}
\usepackage{xlop}
\usepackage{tocloft}
\usepackage{xcolor}
\usepackage[utf8]{inputenc}
\usepackage[a4paper]{geometry}
\usepackage{fancyhdr}
\newcolumntype{M}[1]{>{\centering\arraybackslash}m{#1}}

\renewcommand*{\pagenotesubhead}[1]{}

\makepagenote
\usepackage[arabic , farsi, frenchb , english ]{babel}
\algnewcommand\algorithmicforeach{\textbf{foreach}}
\algnewcommand\ForEach{\item[ \algorithmicforeach]}
\algnewcommand\algorithmicendforeach{\textbf{end}}
\algnewcommand\EndForEach{\item[ \algorithmicendforeach]}
\algnewcommand\algorithmicforeachesp{\textbf{\quad\quad foreach}}
\algnewcommand\ForEachesp{\item[ \algorithmicforeachesp]}
\algnewcommand\algorithmicEndesp{\textbf{\quad\quad end}}
\algnewcommand\Endesp{\item[ \algorithmicEndesp]}
\algnewcommand\algorithmicIff{\textbf{\quad if}}
\algnewcommand\IFf{\item[ \algorithmicIff]}
\algnewcommand\algorithmicIffend{\textbf{\quad end}}
\algnewcommand\IFfend{\item[ \algorithmicIffend]}
\algnewcommand\algorithmicIf{\textbf{\quad\quad\quad if}}
\algnewcommand\IF{\item[ \algorithmicIf]}
\algnewcommand\algorithmicIfe{\textbf{\quad\quad\quad\quad if}}
\algnewcommand\IFe{\item[ \algorithmicIfe]}
\algnewcommand\algorithmicIfee{\textbf{\quad\quad\quad\quad\quad if}}
\algnewcommand\IFee{\item[ \algorithmicIfee]}
\algnewcommand\algorithmicIfeeend{\textbf{\quad\quad\quad\quad\quad end}}
\algnewcommand\IFeeend{\item[ \algorithmicIfeeend]}
\algnewcommand\algorithmicIfend{\textbf{\quad\quad\quad end}}
\algnewcommand\IFend{\item[ \algorithmicIfend]}
\algnewcommand\algorithmicIfeend{\textbf{\quad\quad\quad\quad end}}
\algnewcommand\IFeend{\item[ \algorithmicIfeend]}
\newcounter{example}[section]
\newenvironment{example}[1][]{\refstepcounter{example}\par\medskip
   \noindent \textbf{Example~\theexample. #1} \rmfamily}{\medskip}

\begin{document}

\begin{titlepage}
\begin{center}

\textsc{\Large TUNISIAN REPUBLIC}\\[0.2cm]  \textsc{\Large Ministry of Higher Education}\\[0.2cm] 
\textsc{\Large and Scientific Research} \\[0.2cm] 
 \textsc{\textsl{University Tunis El manar} } \\[0.2cm] 
\includegraphics[width=0.25\textwidth]{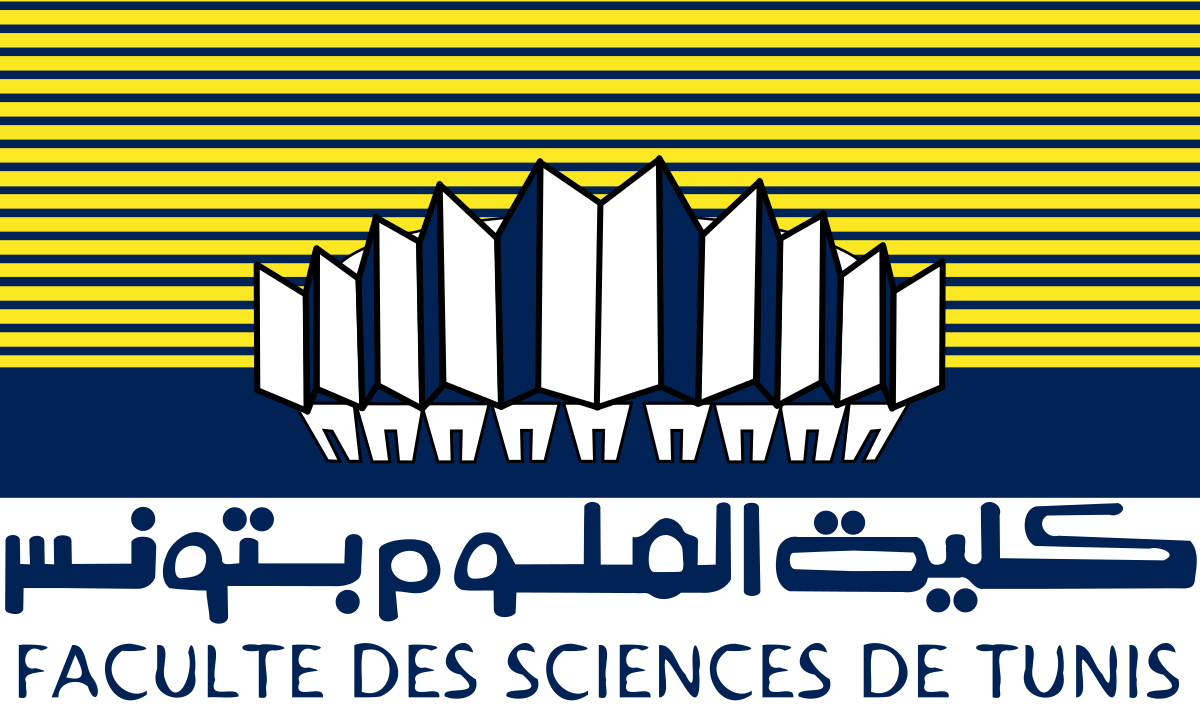}~\\[0.1cm]
\textsc{\Large Faculty of Sciences of Tunis
} \\[0.2cm] 
 \textsc{\Large DEPARTMENT OF COMPUTER SCIENCE}\\[1cm]
{\huge \bfseries Master thesis} \\[1cm] 
\textsc{\Large submitted for a Master's degree in Computer Science}\\[1cm]
\textsc{\Large BY}\\[1cm]

\textbf{ \Large Sihem Sahnoun}\\[1cm]  
\textbf{\textsc{\huge Event extraction based on open information }}\\[0.3cm]
\textbf{\textsc{\huge extraction and ontology}}\\[0.3cm]
\textsc{Defended on January 28, 2019}\\[2cm]
\vfill
\end{center} 

\end{titlepage}
\begin{spacing}{1.3}

\include{remerc}
\chapter*{\centering \textit{Acknowledgements}}
It's with great pleasure that I reserve this page as a sign of gratitude  to all those who helped me during the realization of this work and in an exceptional way:\newline
\newline
\textbf{Mr. Habib Ounelli} the president and \textbf{Mr. Sami zghal} the examiner, and who were my master professors, thanks to the training that they have given to us, we are  grateful for you to reach this level.\newline\newline
\textbf{Mr. Samir Elloumi }, my master thesis director for believing in me and giving me the opportunity to realize this research .
\newline
Thanks to you, I was able to discover a new scientific field which is the event extraction.
I had the privilege of working with you and having a good sense of your qualities and values.
\newline
Your seriousness, your competence and your sense of duty have marked me enormously.
Here you will find the expression of my deep admiration for all your scientific and human qualities.
\newline
\newline
I would also like to thank  warmly:\newline
All members of the IT department for their welcome. And more particularly my colleagues for the daily mood. I thank them for their encouragement and good humor.
\thispagestyle{empty}
\chapter*{\centering\textit{Dedication}}
\vspace{1.5cm}
\textbf{To my very dear mother}, no dedication can  express my  deep gratitude  for her sacrifice,\newline her encouragement and her love she has given me since birth.
May God, preserve her and grant her happiness, health, and a long life.
This work is the fruit of your sacrifices that you have made for my education.\newline\newline
\textbf{To my late father}, who unfortunately didn't stay in this world long enough to see his daughter reached this level.\newline \newline
\textbf{To my dear uncle Noureddine and his wife Najoua.} My uncle who is my father, my friend, a brother, the words are not enough to express the attachment, the love and the affection that I carry for him.\newline\newline
\textbf{To my brothers Kamel, Radhouane and Bassem} who have always encouraged me throughout my studies.
\newline \newline
\textbf{To my sister Senda and her husband Mohamed.} My sister thank you that you are always at my side I love you.\newline\newline
\textbf{To my nieces Nour and Nesrine}, they give to our life a meaning since their birth.\newline \newline
\textbf{To all my friends, for the friendship that has united us and the memories of all the moments we have shared. I dedicate this work to you with all my wishes for happiness, health and success.}
\thispagestyle{empty}
\newpage
\pagenumbering{roman}

\renewcommand{\contentsname}{Table of contents}
\tableofcontents
\thispagestyle{empty}
\newpage

\renewcommand{\listfigurename}{List of Figures}
\listoffigures
\thispagestyle{empty}
\newpage
\renewcommand{\listtablename}{List of Tables}
\listoftables
\thispagestyle{empty}
\newpage
\listofalgorithms
\thispagestyle{empty}
\newpage

\end{spacing}
\begin{spacing}{1.5}
\setcounter{page}{1}
\renewcommand{\chaptername}{Chapter}
\renewcommand*{\chaptername}{Chapter}
\chapter*{General Introduction}
\markboth{General Introduction}{}
\addcontentsline{toc}{chapter}{General Introduction}
Most  of the information  is available in the form of unstructured textual documents due to the growth of information sources (the Web for example).
Information Extraction (IE) is developing as one of the areas of active research in artificial intelligence. It was developed in the late 1980s and 1990s with the Message Understanding Conferences (MUC) \cite{maReference1} in which a set of evaluation campaigns have been suggested and have defined the different tasks of the IE systems.
\newline
 Their purpose was to identify, from the technologies of the components developed for the  information extraction, applications that would be practical and  realized in a short term, which allow computers to read, understand and organize unstructured text into a knowledge base \cite{maReference15,maReference16}.
There are different ways of information extraction, we can cite: Named entity recognition, event extraction, and relationship identification.\newline
\newline
The named entity recognition plays an important role in the IE which allows the identification of PERSON, ORGANIZATION, LOCATION, etc. This recognition has been the subject of several campaigns that have addressed the problem of identifying named entities (NEs) in various fields (medical, political, etc.) and  in various languages (English, French, Arabic, etc.).\newline
\newline
The  event extraction has been also well studied for more than two decades, primarily through the Message Understanding Conferences (MUC) and the Automatic Content Extraction (ACE) programs.\newline
Each year of the MUC program focused on a single type of event template, allowing for the study of complex structure and fillers within the event. Among the types of events studied in the MUC programs are \cite{maReference20}:  Fleet operations(1989),  terrorist activities in Latin America (1992),  corporate joint ventures,\textcolor{white}microelectronic production (1993).\newline
With the ACE program, researchers have switched focus to more general types of events,  such as conflicts, transportation of people/items, and life events. This allows the program to capture a much wider range of event types than the MUC program.
\newline
In 2007, the open information extraction (OIE) has appeared and has allowed the task of extracting knowledge from texts without much supervision. OIE systems aim to obtain relation tuples with a highly scalable extraction by identifying a variety of relation phrases and their arguments in arbitrary sentences.
\newline
The OIE presents a new important challenge for IE systems, including an automatic extraction of relations unlike the traditional IE systems, which use manually labeled inputs \cite{maReference15}.\newline
\newline
Our goal is about an event extraction by using
 an OIE system and an ontology. 
\newline 
This report describes our new approach and is structured as follows:
Chapter 1 looks in more detail at the different types of IE and their related works, and a detailed description of the OIE approaches. This description allows us to have a thorough idea which is strongly related to our research work.
\newline
\newline
Chapter 2 looks in more detail at the description of existing systems in event extraction. this description allows us to see the work done in this area with a brief review of the properties of each work that is presented in a summary table.
\newline
\newline
The technical details of implementing our new approach are described in Chapter 3. The first section of this chapter outlines the system architecture of our approach. The second section describes the algorithms of our approach. The third section is devoted to examples to illustrate the algorithms of the previous section. 
\newline 
\newline At the level of chapter 4, we evaluate our system of event extraction by applying some metrics of tests. \newline 
Finally, we end our research report with a general conclusion that highlights the specific contribution of our research work and to conclude with some perspectives.
\chapter{Information Extraction:  Fundamental  Aspects }
\section{Introduction}
In this chapter, we will begin with a general presentation of IE,  we will focus on the named entity recognition (NER), the relation identification between two entities and the event extraction. We will concentrate on the definition and utility,  and the related works for each  type of IE. Subsequently, we will talk about OIE as well as the main approaches that are related to it.
\section{Information Extraction}
There are different tasks of IE  such as NER which based on the extraction of categorizable textual objects in classes such as names of people, names of organizations, etc. The relation identification is another task in the field of IE which aims to find the mention of a binary relation between two entities in a text. Another specific type of knowledge that can be extracted from the text is the event and can be considered as an object that admits an existence in the time space and depends on other objects \cite{maReference3}. In this section, we present an overview of the IE domain and we explain its different types.
\newpage
\subsection{Named Entity Recognition}
The task of NER is sensitive to the result of natural language processing (NLP). This recognition is the subject of several works because it constitutes an essential element for the extraction of more complex entities.
\newline
\begin{enumerate}[label=\alph*)] 
\item  \textbf{Definition and utility} \newline 
A named entity (NE) is often a word or an expression that represents a specific object of the real world.
The NEs generally correspond to the names of the person, organization, place and  dates, monetary units, percentages, units of measurement, etc.
Therefore NER is defined as an important task among the tasks of IE, which has been satisfactorily carried out for well-formed texts such as news story for certain languages like english and french.
NER consists of searching and identifying objects which are categorizable into two classes \cite{maReference28}:
\newline
\begin{itemize}
\item \textbf{Named entities}  can be the name of people, name of organizations, name of places, etc.\newline The name of people extractor aims to identify the first and the last name, the places extractor based on a list of files that contains the name of places and their types (cities, countries, governorate) and the organization extractor aims to identify the name of organization and its type (Association, society, Faculty, etc).
\newline
\item \textbf{Numeric entities}  can be temporal expressions and numbers.\newline
The temporal expressions can be dates and any other temporal markers (period, date, etc).\newline
Numbers can be numerical expressions such as money and percentages.
\end{itemize}
\newpage
\begin{example}
\textbf {Named entities}\newline
Figure \ref{Example of named entities} illustrates an example of named entity extraction from a text written in english \cite{maReference3}. 
\newline
Purple labels refer to named entities of type ORGANIZATION,  named entities in red are of type POSITION and named entities in blue are of type PERSON (NAME).
\newline
\begin{figure}[ht]
  \begin{center}
    \mbox{\includegraphics[width=15cm, height=3cm]{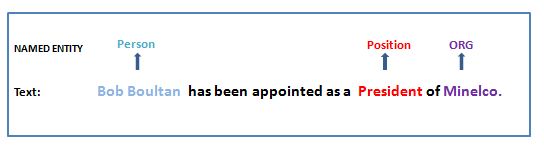}}
      \caption{Example of named entities}
    \label{Example of named entities}
  \end{center}
\end{figure}
\end{example}
\textbf{\item Related works for NER}
\newline
NER is a very active area of research that has received the attention of many researchers and has produced the most fruitful results. There are several NER systems that exist and are classified into three broad families:  Systems based on\textbf{ symbolic approaches}, systems based on \textbf{ learning approaches} and \textbf{hybrid systems}  \cite{maReference2}.
\newline
\newline
\textbf{The symbolic approach}  is based on the use of formal grammars built by hand and  exploit syntactic labeling associated with words, such as the grammatical category of the word.
\newline
It is based also on dictionaries of proper names which usually include a list of the most common names , first names, place names, and names of organizations. \newline 
The work of Shaalan and Raza \cite{maReference29} who developed their NERA system to extract ten types of NEs. This system relies on the use of a set of NE dictionaries and a grammar in the form of regular expressions. We also find the work of Zaghouani \cite{maReference30} who presented a module for locating NEs based on rules for the arabic language.
\newline
\newline
To learn patterns that will recognize entities, \textbf{learning-based methods} use annotated data.\newline
The annotated data corresponds to documents in which the entities, with their types, are indicated. Subsequently, a learning algorithm will automatically develop a knowledge base using several numerical models (CRF, SVM, HMM ...) which is not the case for symbolic approaches that only apply the previously injected rules. Benajiba \& al \cite{maReference31} developed an SVM learning technique for their NER system using a set of features of the arabic language. For poorly endowed languages like Telugu P Srikanth and Kavi Narayana Murthy \cite{maReference33} used the CRF technique to extract named entities.
\newline
\newline
The combination of the two antecedent approaches represents the emergence of an \textbf{hybrid approach}. It uses rules written manually but also builds some of its rules based on syntactic information and information extracted from learning data.
Abuleil \cite{maReference32} has adopted an hybrid approach for extraction of entities in arabic, taking advantage of symbolic and learning approaches.
\newline
\newline
As the recent advancement in the deep learning (DL), it  produces huge differences in accuracy compared to traditional methods for Natural Language Processing (NLP) tasks.  Entity extraction from text is a major Natural NLP task. According to Jason P.C. Chiu \& Eric Nichols (2016) \cite{maReference50}, the implementation has a bidirectional long short-term memory(BLSTM) at its core and a convolutional neural network (CNN) to identify character-level patterns.
\end{enumerate}
\newpage
\subsection{Relation identification}
The extraction of relationships between named entities is an important task for many applications and many studies have been proposed in different frameworks such as  question-and-answer system, extraction of networks, etc \cite{maReference4}.
\newline
\begin{enumerate}[label=\alph*)] 
\item\textbf{Definition and utility} 
\newline
RE is a very useful step in IE. The relation extraction task (RE) involves identifying relationships of interest in each sentence of a given document.
\newline
A relation usually indicates a well-defined relation (having a specific meaning) between two or more NEs. 
 \newline
 \begin{example}\textbf{Example of RE}\newline
We can distinguish two cases of RE \cite{maReference3}:  \newline
 The first concerns the identification of a relationship when both entities are pre-identified in the text,
 the second concerns the identification of all existing relationships between all  entities that can be found in an open corpus (here we are in the case of OIE, we will discuss about it in more detail in the next section).
\newline
 \begin{figure}[ht]
  \begin{center}
    \mbox{\includegraphics[width=15cm, height=4cm]{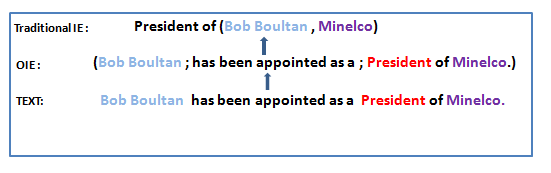}}
      \caption{Example of RE}
    \label{Example of RE}
  \end{center}
\end{figure}
\end{example}
 \newpage
\item \textbf{Related works for RE}
\newline
The existing IE systems can be roughly categorized along two dimensions: The supervision
required and the used models.
\newline
A system can be fully supervised, semi-supervised, or distantly supervised. The second dimension is that the model used can be pattern-based,  not pattern-based (most of them are feature-based sequence classifiers), or an hybrid of pattern and feature-based. 
\newline
\begin{itemize}[label=\textbullet]
\item\textbf{The pattern-based fully supervised approach} \newline 
Systems differ in how they create patterns, learn patterns, and learn the entities they extract. It has been very successful at extracting richer	relations using	
rules and NEs. The  work by Hearst (1992) \cite{maReference35} used hand written rules to automatically generate more rules that were manually evaluated to extract hypernym-hyponym pairs from text.
Berland \& Charniak (1999) \cite{maReference36} used patterns to find all sentences in a corpus containing
basement and building.
\newline
\item \textbf{The pattern-based distantly supervised}
\newline
The matching of seed sets to text take word ambiguity into account when two different objects with the same lexicalisation express two different relations by generating a training data automatically.
For example, the DIPRE algorithm by Brin (1998) \cite{maReference37} used
string-based regular expressions in order to recognize relations such as author-book, while the
SNOWBALL algorithm by Agichtein \& Gravano (2000) \cite{maReference38} learned similar regular expression patterns
over words and named entity tags.
\newline
\item\textbf{Distantly supervised Non-pattern-based systems} \newline
Many systems, such as CRFs-based systems, use\textcolor{white}a{0}set of entities, dictionaries as features. Cohen \& Sarawagi
(2004) \cite{maReference39} worked on improving the matching of named entity segments to dictionaries and to use it as a feature for a sequence model.
\newpage
\item\textbf{Distantly supervised hybrid systems}\newline
Individual components of a pattern-based learning system can use feature-based learning methods to learn good pattern and entity ranking functions.  DeepDive (Niu \& al., 2012) \cite{maReference40} has shown promising results on distantly supervised relation extraction of (Angeli et al., 2014) \cite{maReference41} by using fast inference in Markov logic
networks. (Angeli \& al. 2014) \cite{maReference41} used learned dependency
patterns.
\newline
\item\textbf{Deep learning
models for relation extraction.}\newline
The word embeddings (Mikolov et al., 2013) \cite{maReference48}  express each word as a vector and  aim to capture the syntactic
and semantic information about the word. They
are learnt using unsupervised methods over large
unlabeled text corpora which were adapted as standard in all
subsequent RE deep learning models. (Nguyen, T. H., \& Grishman, R. 2015) \cite{maReference49},  explore CNNs for  Relation Extraction and Relation Classification tasks, the model completely gets rid of
exterior lexical features to enrich the representation
of the input sentence and lets the CNN learn
the required features itself.
\newline
\item\textbf{Open information extraction}\newline
OIE, a popular task in recent years, it extracts relations	from the	web	with no	training	data and no	list of	relations. ReVerb (Fader \& al., 2011) \cite{maReference11} and OLLIE (Mausam \& al., 2012)  \cite{maReference5} learn how to extract open-domain relation triples. The section 2.3 describes many systems in more details.
\end{itemize}
\end{enumerate}
\newpage
\subsection{Event recognition}
The event detection task is the most fundamental at the IE level where we have retained some definitions based on its utility and the techniques used to identify the events.
\newline
\newline
\begin{enumerate}[label=\alph*)] 
\item\textbf{Definition and utility}
\newline
Event extraction is intended to extract from the text a characterization of an event, defined by a set of entities associated with a specific role in the event.
\newline
The extraction of events from unstructured data could be beneficial in various ways for example, the news messages can be selected more accurately, depending on the preferences of the user and the topics identified,  economic events such as mergers and acquisitions, play a crucial role in the day-to-day decisions, etc \cite{maReference8}. \newline
\begin{example}
\textbf{Two levels event extraction
 \cite{maReference43}}\newline
Figure \ref{Example of two levels event extraction}  highlights an example of an event extraction related to the management change. This example was made on a text extracted from a press release. The event is considered to be an action allowing the management change and the characteristics of the event are: dates, people, places, etc.
\newline
\begin{figure}[ht]
  \begin{center}
    \mbox{\includegraphics[width=15cm, height=4cm]{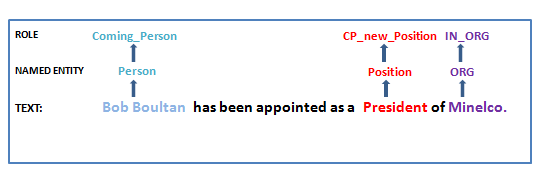}}
      \caption{Example of two levels of event extraction}
    \label{Example of two levels event extraction}
  \end{center}
\end{figure}
\end{example}
\newpage
\item\textbf{Related works for Event recognition }
\newline
Some techniques use \textbf{data-driven approaches}, others use \textbf{knowledge-driven approaches} and others use \textbf{hybrid approaches}.
\begin{itemize}[label=\textbullet]
\item\textbf{Data-driven approaches} require a large text corpora in order to develop
models that approximate linguistic phenomena.  Data-driven methods require a lot of data and a little domain knowledge and expertise,
while having a low interpretability. They don't deal with meaning explicitly, i.e., they discover relations in corpora without considering semantics.\newline 
For example Conditional Random Field based (CRF) \cite{maReference44} systems apply the classifier to a set of texts to produce a set of annotated texts.
The interest and efficiency of CRFs come from taking into account the dependencies between labels related to each other in the graph.
\newline
\item\textbf{knowledge-driven approaches} is often based on models that express rules representing expert knowledge.
 It is intrinsically based on linguistic and lexicographical knowledge, as well as on existing human knowledge concerning the content of the text to be treated. This alleviates the problems with the statistical methods concerning the meaning of the text. For example the GLAEE approach \cite{maReference3} is based on the generation of annotation patterns that involves a list of keywords and cue words which purpose  to identify events in the learning phase,  afterwards the annotation phase is performed by an alignment between the pattern and the new text.
\newline
\item \textbf{hybrid approaches} seem to be a compromise
between data and knowledge-based approaches, requiring an average amount of data and domain knowledge and having medium interpretability. However, it should be noted that the amount of expertise required is high, due to the fact that several techniques are combined. For example the interest of Two-level approach \cite{maReference43} is to adapt the recognition of named entities  level  to the CRF tool based on learning techniques and a correspondence between level 1 learning (PERSON, ORG, DATE, NUMEX, PROFIL) as well as learning level 2 (NEW PERSON, COMING PERSON) which brings us back to a double generation of the classifier.
\end{itemize}
To overcome the difficulties of complicated feature engineering
and domain dependency, researchers used neural network approach
for event extraction \cite{maReference51,maReference52,maReference53}.  All these works deal with
english language and principle objective of these tasks is to detect
the trigger word in the text which indicate an event.
\end{enumerate}
\section{Open information extraction}
In traditional IE systems, the major goal was to extract some specific relations in a set of hand-labeled documents but there are two fundamental problems with these IE systems that prevent us to use it in IE tasks nowadays. First, we prefer to extract all of the existing relations in a set of documents automatically rather than looking for pre-specified relations. Second, traditional IE systems should have hand-labeled documents which are on a specific topic. Hand labeling documents is somehow impractical nowadays because of the huge number of documents present, for example all of the webpages. Therefore, there has been a need towards the next generation of IE systems.
\newline
In 2007, an approach was introduced from the
RE task, called OIE, which scales RE on the Web. In the OIE task, the relation names are not known in advance. The only entry of an OIE system is a corpus. The table 1.1 summarizes the differences between traditional IE systems and OIE \cite{maReference10}.
\newline
\begin{table}[!ht]
\centering
\begin{tabular}{|p{3cm}|p{3cm}|p{3cm}|  }
\hline
& \textbf{Traditional IE}&\textbf{OIE} \\
\hline
\textbf{Input} & corpus + hand-labeled data &corpus \\
\hline
\textbf{Relations} & specified in advance   & discovered automatically\\
\hline
\textbf{Extractor} & specific relation &independent relation  \\
\hline
\end{tabular}
\caption{Difference between Traditional IE and OIE}
\label{table:1}
\end{table}
\newline
\newline
OIE is currently being developed in its second generation in systems such  as ReVerb \cite{maReference11}, OLLIE
\cite{maReference11}, and ClausIE \cite{maReference12}, which extend from Open IE previous
systems such as TextRunner \cite{maReference13} and
WOE \cite{maReference14}. In this section  we will describe an overview of two generations of open IE systems \cite{maReference15}.
\subsection{First Open IE generation}
Several OIE systems have been proposed in the first generation, including TextRunner \cite{maReference13}, WOE \cite{maReference14} that researchers often focus in their research.
\begin{enumerate}[label=\alph*)] 
\item\textbf{TextRunner \cite{maReference13}}\newline
TextRunner is one of the first OIE systems developed by the researchers from the University of Washington. The architecture of the system is displayed below in Figure \ref{Architecture of TextRunner}.
\begin{figure}[htt]
  \begin{center}            \mbox{\includegraphics[width=18cm, height=9cm]{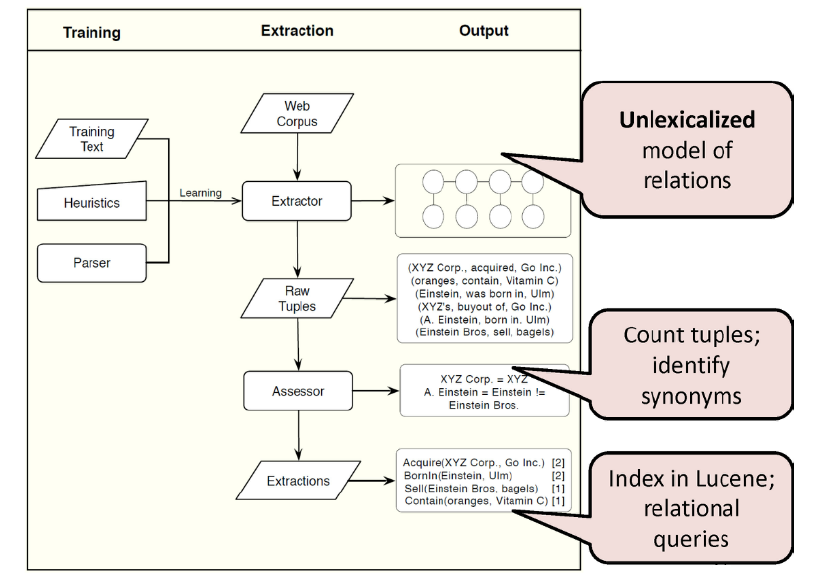}}
      \caption{Architecture of TextRunner }
    \label{Architecture of TextRunner}
  \end{center}
\end{figure}
\newline
The goal of this system is to extract some tuples from the data which are in the form of (Arg1,Rel,Arg2). "Arg1" and "Arg2" are two noun phrases, and "Rel" is the relationship between these two arguments. One important assumption in this model is the redundant information, so that, this redundancy and co-occurrence of words can be used to determine the correctness of a tuple.
\newline
This system generally consist of the following three steps: 
Intermediate Levels of Analysis, Learning Models and Presentation\cite{maReference15}.
\newline
\begin{itemize}[label=\textbullet]
\item\textbf{Intermediate levels of analysis}
\newline
Entity Recognition (NER), part of speech (POS) and Phrase-chunking are NLP techniques that are used during this stage. A sequence of words are taken as input and each word is assigned in accordance with its syntactic functions e.g., noun, verb, adjective by a POS tagger. A phrase chunked divides phrases in the  sentence based on POS tags. NER identifies and categorizes named entities in a sentence. Some systems such as TextRunner \cite{maReference13}, establish relationships between the head token and its child token by a dependency parser as shown in Figure \ref{POS, NER and DP analysis}. 
\newline
\begin{figure}[ht]
  \begin{center}
 \mbox{\includegraphics[width=13cm, height=6cm]{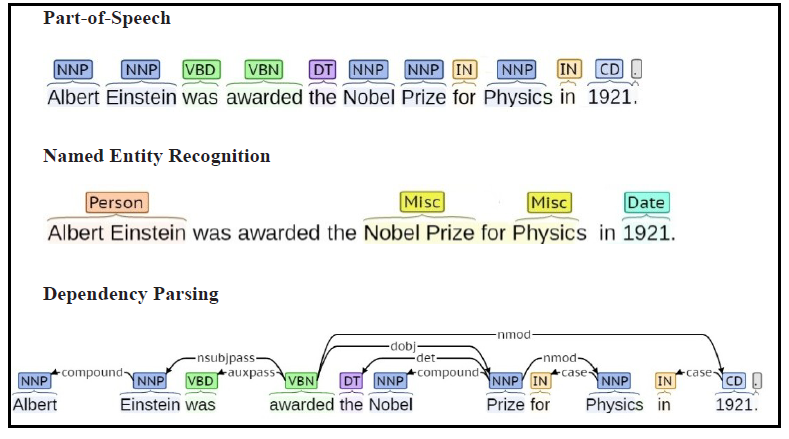}}
      \caption{POS, NER and DP analysis}
    \label{POS, NER and DP analysis}
  \end{center}
\end{figure}
\item\textbf{Learning models} 
\newline
The role of \textbf{ the extractor} is to find the raw tuples from the data. Extractor uses various learning methods like  CRFs to learn if sequences of tokens are part of a relation. CRFs are based on a graphical model that defines the dependencies between random variables by a graph. The main interest of working with an independence graph is that it allows to look for the most probable annotation. It also uses parsers and some heuristics to find the main noun phrases in a sentence and the relationship between them. When identifying entities, the system determines a maximum number of words and their surrounding pair of entities which could be considered as possible evidence of a relation. Figure \ref{Example of a  CRF} shows an example of a CRF to find the main noun phrases in a sentence and the relationship between them.
\newline
\begin{figure}[htt]
  \begin{center}
 \mbox{\includegraphics[width=10cm, height=
    4cm]{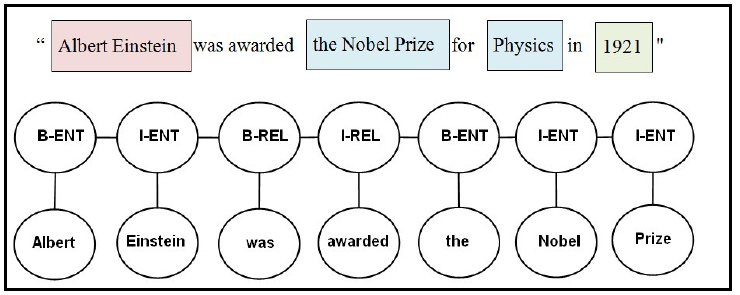}}
      \caption{Example of a  CRF}
    \label{Example of a  CRF}
  \end{center}
\end{figure}
\item\textbf{Presentation}
\newline
\textbf{The assessor}  does some inference over the extracted tuples. For instance, if two entries (noun phrases)   have similar tuples, we can say that with a high probability these entries  are the same.  As an example of this kind of inference, TextRunner may find that “A. Einstein” is the same as “Albert Einstein” and these are not the same as “Einstein Bros.” which is a corporation. Also, if two relations share the same tuples (their entries are similar), again we can say that with a high probability these two relations mean the same thing. The next part is to assign a likelihood probability with a certain tuple using the number of “distinct” sentences in the argument.
\end{itemize}
\newpage
\item\textbf{Wikipedia-based Open IE (WOE) \cite{maReference14}}
\newline
The goal of WOE is to learn an open extractor without direct supervision, the inputs of this system are:  Wikipedia (as source for sentences) and DBpedia (as source for
cleaned infoboxes).
\newline
The key idea underlying WOE is the automatic
construction of training examples by heuristically matching Wikipedia infobox values and corresponding
text.
\newline
These examples are used to generate an unlexicalized, relation-independent (open) extractor. \newline
As shown in Figure \ref{Architecture of WOE}, WOE has three main
components: Preprocessor, matcher, and learner.
\newline
\begin{figure}[htt]
  \begin{center}
    \mbox{\includegraphics[width=10cm, height=5cm]{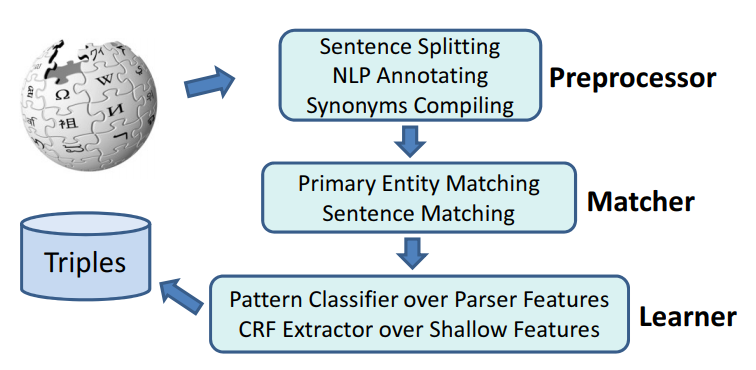}}
      \caption{Architecture of WOE }
    \label{Architecture of WOE}
  \end{center}
\end{figure}
\begin{itemize}[label=\textbullet]
\item\textbf{The preprocessor} 
\newline
The preprocessor converts the raw Wikipedia text
into a sequence of sentences, attaches NLP annotations,
and builds synonym sets for key entities.
\newline
It first renders
each Wikipedia article into HTML, then splits the
article into sentences using OpenNLP.\newline
The preprocessor uses OpenNLP to supply POS tags and NP-chunk annotations
or uses the Stanford Parser to create a
dependency parse.
\newline
It compiles synonymies to increase the recall of the matcher.
This is useful because  Wikipedia articles contain different
mentions of same entities (across pages and between infobox and Wikipedia page).
The preprocessor uses  Wikipedia redirection pages and backward
links to construct automatically synonym sets.
\item\textbf{The matcher}
\newline
The matcher constructs training data for the
learner component by heuristically matching
attribute-value pairs from Wikipedia articles containing
infoboxes with corresponding sentences in
the article.  It Iterates through all its attributes looking for a unique
sentence that contains references to  the subject of the article and the
attribute value (or its synonym).
\newline
Given the article on “ Stanford University ”.\newline
For example, the matcher should associate: \newline
\emph{<established, 1891>} with the sentence \emph{“ The
university was founded in 1891 by... ”}
\newline
$\rightarrow$ \emph{<arg1=Stanford University  ,rel=???,  arg2=1891>} \newline
\item \textbf{The learner}\newline
The learner acquires the open
extractors using either
parser features or POS
features. The first step is the \textbf{construction of relation in the text} by selecting all tokens of expanded path as value for relation ("was not
born" in our example as shown in figure \ref{The parser features used in the learner } ). \newline
The second step is the \textbf{generalization of patterns} by
ignoring CorePaths that don‘t start with subject like
dependencies (s.a. nsubj, nsubj-pass)
 and generalizing CorePaths with substituting lexical words by their POS: Map all noun tags to N, verb tags to V, prep tags to prep, etc. As shown in figure \ref{The parser features used in the learner }.\newline 
 \newpage
\emph{PS: The \textbf{ExpandPath} based on adding all adverbial and adjectival modifiers
"neg" and "auxpass" labels of the root node.}\newline
\emph{The \textbf{CorePath} is the shortest dependency path between arg1 and arg2.}
\begin{figure}[htt]
  \begin{center}
    \mbox{\includegraphics[width=8cm, height=7cm]{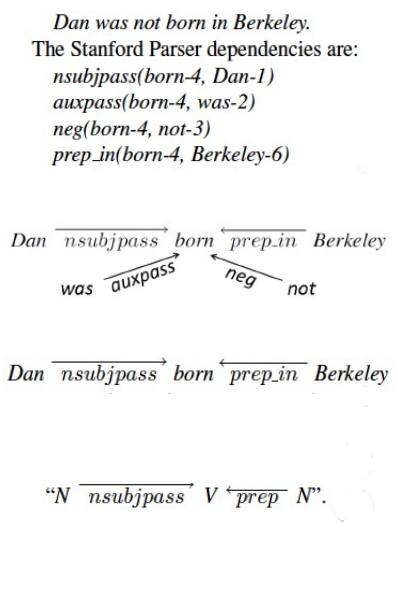}}
      \caption{The parser features used in the learner }
    \label{The parser features used in the learner }
  \end{center}
\end{figure}
\end{itemize}
\textbf{OIE systems of first generation } can suffer from problems such as the extraction of incoherent and non-informative relations. See tables 1.2 and 1.3.
\newline
\textbf{TextRunner and WOE}  don't extract the complete relation between two noun phrases, and extract only part of the relation which is ambiguous.
\newline For instance, where it should extract the relation "is an album by", it only extracts "is" as the relation.
\begin{table}[ht!]
\centering
\begin{tabular}{|p{4cm}|p{4cm}|  }
\hline
Sentences& Incoherent Relation  \\
\hline
The guide contains dead links and omits sites &contains omits\\
\hline
The Mark 14 was central to the torpedo scandal of the fleet& was central torpedo
\\
\hline
They recalled that Nungesser began his career as a precinct leader & recalled began\\
\hline
\end{tabular}
\caption{Example of incoherent extractions}
\label{table:2}
\end{table}
\begin{table}[ht!]
\centering
\begin{tabular}{|p{1cm}|p{7cm}|}
\hline
is& is an album by, is the author of, is a city in  \\
\hline
has & has a population of, has a ph.D.in , has a cameo in\\
\hline
made& made a deal with, made a promise to
\\
\hline
took & took place in, took control over,took advantage of\\
\hline
gave & gave birth to, gave a talk at,gave new meaning to\\
\hline
got & got tickets to, got a deal on , got funding from \\
\hline
\end{tabular}
\caption{Example of uninformative relations(left) and their completions(right)}
\label{table:3}
\end{table}
\end{enumerate}
\newpage
\subsection{Second Open IE generation}
The second generation systems such as REVERB, OLLIE \cite{maReference5}, and ClauseIE \cite{maReference21} deal with inconsistent and non-informative extractions that occur in the first generation by identifying a more meaningful relationship expression.
Several OIE systems have been proposed after TextRunner and WOE, including ReVerb, OLLIE,  ClausIE with two extraction paradigms, namely verb-based relation  extraction and clause-based relation extraction.
\newpage
\begin{enumerate}[label=\alph*)] 
\item\textbf{ ReVerb\cite{maReference5,maReference11}}
\newline
ReVerb,
which implements a general model of verb-based relation
phrases, expressed as two simple constraints:  \textbf{Syntactic and lexical constraint.}
\newline 
\begin{itemize}
\item To satisfy \textbf{the syntactic constraint} three grammatical structures are considered for relations as shown in figure \ref{Three grammatical structures in ReVerb}.
\begin{figure}[htt]
  \begin{center}
    \mbox{\includegraphics[width=8cm, height=4cm]{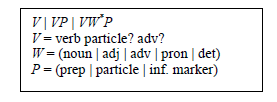}}
      \caption{Three grammatical structures in ReVerb}
    \label{Three grammatical structures in ReVerb}
  \end{center}
\end{figure}
\newline
The pattern limits the
relation phrases to be either a simple verb phrase (e.g., invented),
a verb phrase followed immediately by a preposition
or particle (e.g., located in), or a verb phrase followed by a
simple noun phrase and ending by a preposition or particle
(e.g., has atomic weight of). If there are multiple possible
matches in a sentence for a single verb, the longest possible
match is chosen.
\newline
\item Consider the sentence
"The Obama administration is offering only modest greenhouse
gas reduction targets at the conference."
The POS pattern will match the phrase:
"is offering only modest greenhouse gas reduction targets at ".
\newline
This phrase satisfies the syntactic constraint,
but this is not a useful relation.\newline
To overcome this limitation, we introduce a \textbf {lexical constraint}.
 It is based on the
use of a large dictionary of relation phrases that are known
to take many distinct arguments. Valid relational phrases should take $\geq$  20 distinct
argument pairs over a large corpus (500M sentences).
In order to allow for minor variations in relation phrases, each relation phrase was normalized by removing inflection, auxiliary
verbs, adjectives, and adverbs.
\end{itemize}
\item\textbf{OLLIE \cite{maReference5,maReference11}}
\newline
OLLIE is another OIE system that extracts verbal relationships. Mausam \& al \cite{maReference5} presented OLLIE as an extended ReVerb system, which means Open Language Learning for IE. OLLIE performs a thorough analysis of the verb-phrase relationship identified then the system extracts all mediated relationships by verbs, nouns, adjectives and others as shown in Figure \ref{OLLIE extraction}.
\begin{figure}[htt]
  \begin{center}
    \mbox{\includegraphics[width=8cm, height=3cm]{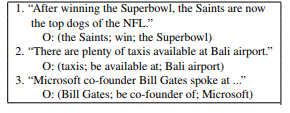}}
      \caption{OLLIE extraction }
    \label{OLLIE extraction}
  \end{center}
\end{figure}
\newline
OLLIE produces a strong performance by extracting relationships not only mediated by verbs, but also mediated by nouns and adjectives.\footnote{https://github.com/knowitall/ollie}
\newline
It can capture N-ary extractions  where the relation phrase only differs by the preposition.
\begin{example}\textbf{N-ary Extraction}\newline
\textbf{Sentence:} I learned that the 2012 Sasquatch music festival is scheduled for May 25th until May 28th.
\newline
\textbf{Extraction:} (the 2012 Sasquatch music festival; is scheduled for; May 25th) 
\newline
\textbf{Extraction:} (the 2012 Sasquatch music festival; is scheduled until; May 28th)
\newline 
\textbf{N-ary:} (the 2012 Sasquatch music festival; is scheduled; [for May 25th; to May 28th])
\newline
\end{example}
OLLIE captures also enabling conditions and attributions if they are present.
\begin{example}\textbf{Enabling conditions}  \newline
\textbf{Sentence:} If I slept past noon, I'd be late for work.
\newline
\textbf{Extraction:} (I; 'd be late for; work)[enabler=If I slept past noon]
\end{example}
\newline
\begin{example}\textbf{Attribution} 
\newline
\textbf{Sentence:} Some people say Barack Obama was not born in the United States.
\newline
\textbf{Extraction:} (Barack Obama; was not born in; the United States)[attrib=Some people  say].
\end{example}
Figure \ref{General architecture of OLLIE} shows the general architecture of OLLIE. The basic idea behind each of the building blocks of OLLIE will be described next.
\begin{figure}[htt]
  \begin{center}
    \mbox{\includegraphics[width=10cm, height=7cm]{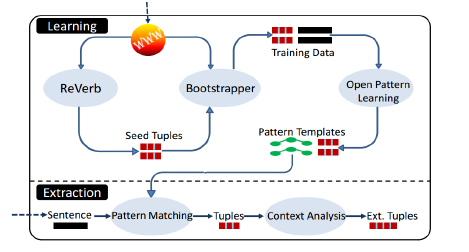}}
      \caption{General architecture of OLLIE}
    \label{General architecture of OLLIE}
  \end{center}
\end{figure}
\newline
OLLIE tries to learn the patterns of relations using their equivalent 
relation found by\textbf{ ReVerb}. In the first step, ReVerb is applied on \textbf{the  raw 
data (WWW)} and among the extracted tuples the most reliable ones are chosen. For each tuple found using \textbf{ ReVerb}. \textbf{The bootstrapper} looks for all the sentences in raw data having the same words as in that extracted tuple.  The next step would be to learn the new patterns of relations using these 
sentences for each of the ReVerb’s extracted tuples. The extracted sentences for each of the ReVerb’s tuples are the \textbf{training data} marked. Using the dependency parsers,  \textbf{Open Pattern Learning} learns and specifies new grammatical formats for relations using the training data which are called \textbf{pattern templates}. For each sentence, we use the output of a dependency parser on the sentence to match it with one or multiple learned pattern templates by \textbf{the Pattern Matching}. Again, using the output of a dependency parser on the sentence, OLLIE tries to change some of the incomplete tuples such that the information within them exactly reflects what was meant by the sentence. For instance, if there is a if clause in the sentence \textbf{the context analysis} block adds it as a ClauseModifier to the tuple.
\item\textbf{ ClauseIE \cite{ maReference8,maReference12}}
\newline
\newline
Another OIE system named ClausIE presented by Corro \& Gemulla [8] uses
clause structures to extract relationships and their arguments from a text. This system obtains and exploits relationship extraction clauses  \cite{maReference14,maReference15}.
\newline
Particularly, in these systems a clause can consist of
different components such as subject (S), verb (V),  object (O), complement (C), and/or one or
more adverbials (A). As illustrated in Table 1.4, a clause can
be categorized into different types based on its constituent.
\newline
\begin{itemize}[label=\textbullet]
\item\textbf{Step 1. Determining the set of clauses} 
\newline
This step seeks
to identify the clauses of the input sentence by obtaining the
head words of all the constituents of every clause. The
mapping of syntactic and dependency parsing are utilized to
identify various clause constituents. Subsequently, a clause
is constructed for every subject dependency, dependent
constitutes of the subject and the verb.
\newline
\newline
\item\textbf{Step 2. Identifying clause types}\newline
When a clause is obtained, it needs to be associated with one of the main
clause types as shown in Table 1.4.  These systems use a decision tree to identify the
different clause types. In this process, the system marks all
optional adverbials after the clause types have been
identified.
\newpage
\item\textbf{Step 3. Extracting relations}
\newline
The systems extract
relations from a clause based on a set of patterns  as illustrated in Table 1.4.
For instance, for the
clause type SV in Table 1.4, the subject presentation “Albert
Einstein” of the clause is used to construct the proposition
with the following potential patterns: SV, SVA, and SVAA.
Dependency parsing is used to forge a connection  between
the different parts of the pattern. As a final step, n-ary facts
are extracted by placing the subject first followed by the
verb or the verb with its constituents. This is followed by
the extraction of all the constituents following the verb in
the order in which they appear. As a result, these systems
link all arguments in the propositions in order to extract
triple relations.
\end{itemize}
\begin{table}[ht!]
\centering
\begin{tabular}{|p{1cm}|p{3cm}|p{1cm}|p{4.5cm}|  }
\hline
Clause types& Sentences& Patterns &Derived clauses \\
\hline
SV &Albert Einstein died in Princeton in
1955. &SV
SVA
SVA
SVAA&(Albert Einstein, died)\newline
(Albert Einstein, died in, Princeton)\newline
(Albert Einstein, died in, 1955)\newline
(Albert Einstein, died in, 1955, [in] Princeton) \\
\hline
SVA & Albert Einstein remained in Princeton
until his death.& SVA
SVAA &(Albert Einstein, remained in, Princeton)\newline
(Albert Einstein, remained in, Princeton, until his death)\\
\hline
SVC & Albert Einstein is a scientist of the 20th
century. &SVC
SVCA&(Albert Einstein, is, a scientist)\newline
(Albert Einstein, is, a scientist, of the 20th century)  \\
\hline
SVO & Albert Einstein has won the Nobel Prize
in 1921. &SVO
SVOA&(Albert Einstein, has won, the Nobel Prize)\newline
(Albert Einstein, has won, the Nobel Prize, in 1921)  \\
\hline
SVOO & RSAS gave Albert Einstein the Nobel Prize. &SVOA&(The doorman, showed, Albert Einstein, to his office)  \\
\hline
SVOA &Albert Einstein declared the meeting
open. &SVOC&(Albert Einstein, declared, the meeting, open)  \\
\hline
\end{tabular}
\caption{Clause types; S: Subject, V: Verb, A: Adverbial, C: Complement, O: Object}
\label{table:4}
\end{table}
Among the restrictions in ClauseIE we can mention that ClauseIE incorrectly identifies ` there 'as the subject of a relation.\newline
\textbf{For example}:" In today's meeting, there were four CEOs ".
\end{enumerate}
\newpage
\section{Conclusion}
In this chapter we have listed the different levels of the IE and their related works, then we examined the part of OIE and how the different systems work.
In the next chapter, we will explain multiple existing approaches for event extraction.
\chapter{Event extraction: State of the art}
\section{Introduction}
Several systems have been proposed in the context of IE as an essential module in different applications. Therefore, it has favored the development of many approaches. In this chapter we represent the main approaches that are strongly related to our work as well as some used tools. We close this chapter by a summary table.
\section{GLAEE}
GLAEE (General learning approach for event extraction)\cite{maReference3} is an approach based on the generation of annotation patterns which involves a list of keywords and cue words.
It purposes to identify events by an alignment between the pattern and the new text.
\subsection{Architecture}
\begin{figure}[htt]
  \begin{center}
    \mbox{\includegraphics[width=15cm, height=10cm]{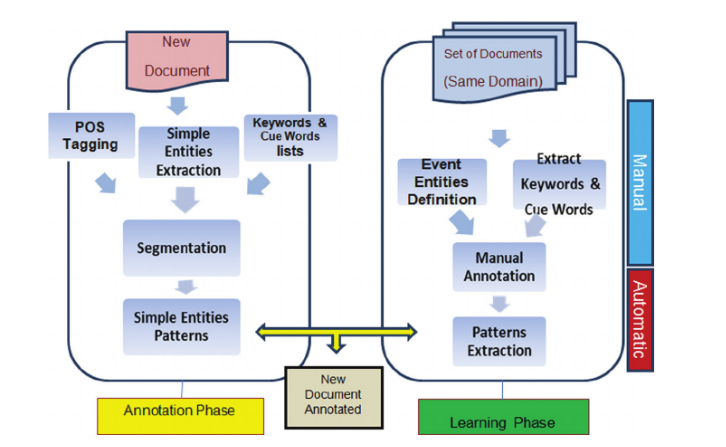}}
      \caption{GLAEE architecture}
    \label{GLAEE  architecture}
  \end{center}
\end{figure}

We consider the following two phases: Learning phase and annotation phase as shown in Figure \ref{GLAEE  architecture}.
\newline
 The main objective of the learning phase is to build a set of patterns describing the possible occurrences of the event. This pattern is generated relatively to the manual annotation. The second phase contains the most important information in the text of entities, keywords and cue words to reach finally the alignment process to identify the possible roles of some entities.
\newline
\subsection{Principle}
\textbf{The extraction of keywords and cue words}
\newline
The  extraction process is a first step of the system that consists in extracting manually the keywords and the cue words from a set of texts and assigning a code to the elements having the same meaning. For example in  Figure \ref{Keywords, cue words and entities encoding}, the keyword "Appoints" and the keyword "Names" are coded by the same code K1. 
\begin{figure}[htt]
  \begin{center}
    \mbox{\includegraphics[width=15cm, height=3cm]{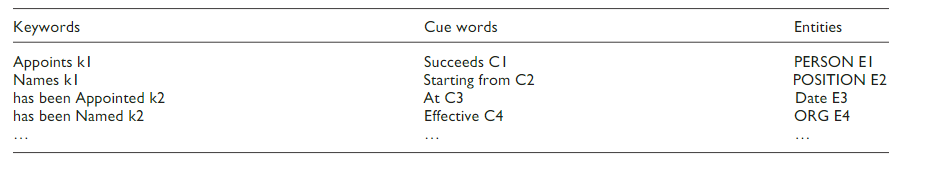}}
      \caption{Keywords, cue words and entities encoding}
    \label{Keywords, cue words and entities encoding}
  \end{center}
\end{figure}
\newpage
\textbf{The manual annotation }\newline
For each text in the learning set,  its associated manual annotation as shown in Figure \ref{Sample of a learning text and its associated manual annotation}.
\begin{figure}[htt]
  \begin{center}
    \mbox{\includegraphics[width=18cm, height=6cm]{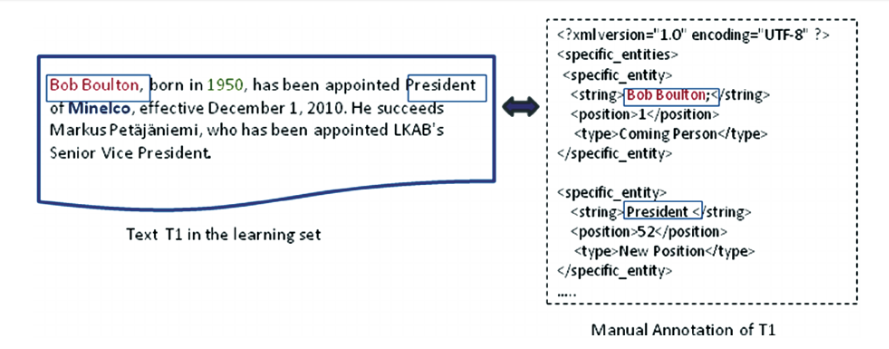}}
      \caption{Sample of a learning text and its associated manual annotation}
    \label{Sample of a learning text and its associated manual annotation}
  \end{center}
\end{figure}
\newpage
This generation has made it possible to extract a coded segment that contains the most important information cited in the text in terms of keywords, cue words and entities, with their associated annotations. Figure \ref{Encoded pattern} describes an example of a pattern.
\begin{figure}[htt]
  \begin{center}
    \mbox{\includegraphics[width=18cm, height=5cm]{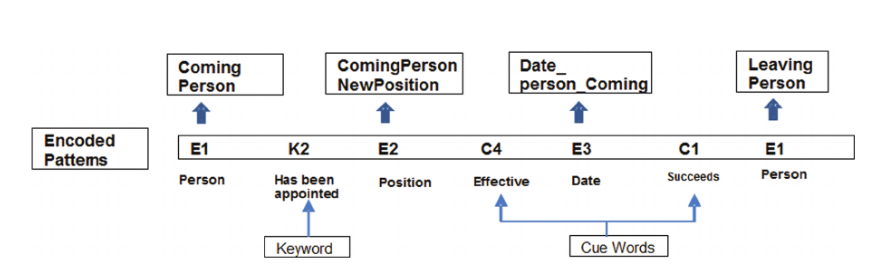}}
      \caption{Encoded pattern}
    \label{Encoded pattern}
  \end{center}
\end{figure}
\newline
\textbf{The alignment phase}
\newline
The aim of the alignment phase is to identify possible roles of certain entities as event description. For example, the alignment between the pattern in Figure \ref{Encoded pattern} and a new text as shown in Figure  \ref{Event extraction for a new text after alignment step}.
\begin{figure}[htt]
  \begin{center}
    \mbox{\includegraphics[width=18cm, height=5cm]{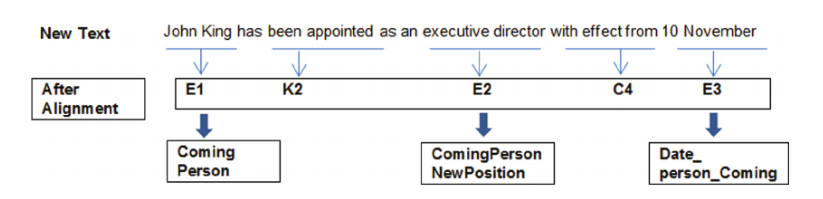}}
      \caption{Event extraction for a new text after alignment step}
    \label{Event extraction for a new text after alignment step}
  \end{center}
\end{figure}
\newpage
\section{CRF(Conditional Random Fields)}
 CRFs \cite{maReference44} are based on a conditional approach to label and tokenize data sequences for representing
a conditional model P(y|x), where both x and y have a non-trivial structure (often
sequential).
\subsection{Architecture} 
\begin{figure}[htt]
  \begin{center}
    \mbox{\includegraphics[width=16cm, height=11cm]{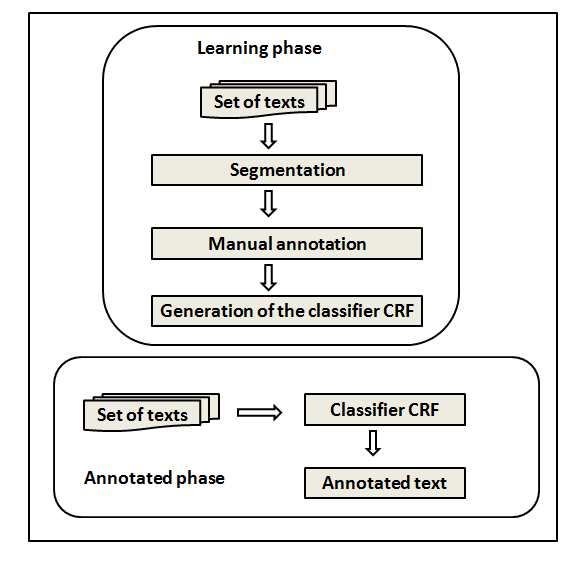}}
      \caption{The CRF architecture}
    \label{The CRF architecture}
  \end{center}
\end{figure}

The CRFs' architecture is presented in Figure \ref{The CRF architecture}. This architecture admits two complementary phases.\newline
The first part is a sequence of steps for the learning process, ie the generation of the CRF classifier.\newline
A second step in which the system applies the classifier to a set of texts to have as a result a set of annotated texts.
\newline
\newpage
\subsection{Principle}
\textbf{Learning phase of CRFs}
\newline
\newline
The mechanism of CRFs is to manually prepare a set of corpora to train the model. This task based on the division of the text to sentences (segmentation), then associate with each word of the sentence a syntactic category. \newline
\newline
\textbf{Annotation phase of CRFs}
\newline
\newline
CRFs model has been introduced as a high-performance alternative to Hidden Markov Models (HMMs), they are often used for object recognition.
It belongs to the sequence modeling family. They are defined by X and Y, two random variables describing respectively each unit of the observation x and its annotation y, and by a graph G = (V, E) such that V = X $\cup$ Y is the set of nodes, \begin{figure}[htt]
 E $\subseteq$  V \mbox{\includegraphics[width=0.3cm, height=0.3cm]{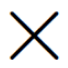}} V is the set of arcs and  p (y | x) is a probability distribution.
\end{figure} 
\newline
This phase makes it possible to look for the most probable annotation by finding the annotation y and maximizing the probability p (y | x).
\newline
The interest and the efficiency of CRFs comes from taking into account the dependencies between labels related to each other in the graph. By looking for the best y, i.e. the best sequence of labels associated with a complete data x.
\newpage
\begin{example}\textbf{CRF}
\newline
Figure \ref{CRF example} illustrates the mechanism of the CRF model for the annotation phase.
This graph describes examples of annotated sentences with their probability distributions.\newline
The interest of this graph is to find the best annotation associated with a new text
\emph{x= "The white lie"} 
according to the CRFs, we obtain the following annotation:\newline
The $\mapsto$ Det, white$\mapsto$ Adj,  lie $\mapsto$ Noun
\newline
\newline
\begin{figure}[htt]
  \begin{center}
    \mbox{\includegraphics[width=15cm, height=10cm]{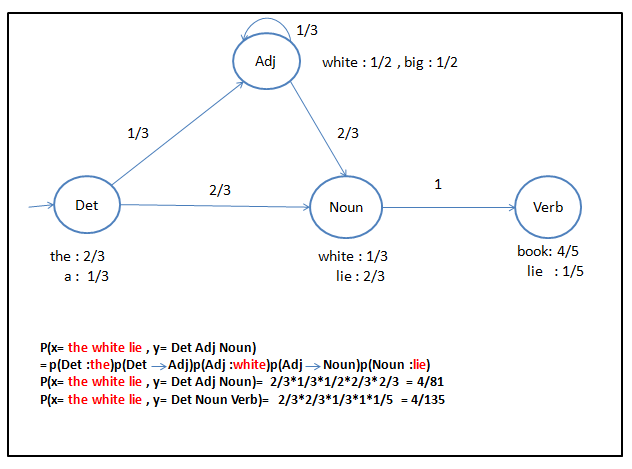}}
      \caption{CRF example}
    \label{CRF example}
  \end{center}
\end{figure}
\end{example}
\newpage
\section{Template filling through information extraction}
This event extraction approach \cite{maReference54} consists of filling  the forms with extracted information  that designates a set of events.
\subsection{Architecture} 
\begin{figure}[htt]
  \begin{center}
    \mbox{\includegraphics[width=15cm, height=10cm]{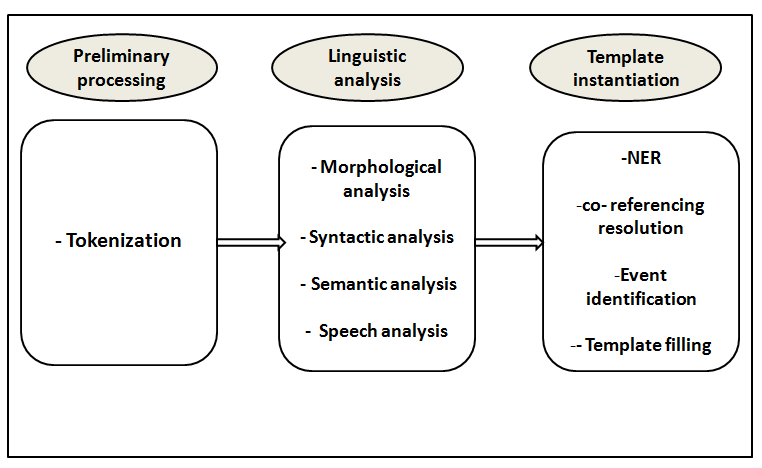}}
      \caption{Template filling system architecture}
    \label{Template filling system architecture}
  \end{center}
\end{figure}

Figure \ref{Template filling system architecture} illustrates the different stages of extracting events in order to complete the template.
\newline
\newpage
\subsection{Principle}
In order to propose an automatic form filling system, this approach proposes a process which admits 3 phases:
\newline
\subsubsection{Preliminary processing}
It consists of breaking up the text into sentences and each sentence into words.
\subsubsection{linguistic analysis phase}
This phase is primarily concerned with a \textbf{morphological analysis} which observes and analyzes the different parts of a word. The root is the origin of the word; when we remove all affixes. Words are composed of units of meaning; the smallest units of meaning are the morphemes. Subsequently, \textbf{the syntactical analysis} consists in producing the grammatical relations: subject-verb or verb-verb-complement relations, prepositional attachments.\newline
Moreover, the \textbf{semantic analysis}  aims to build from each proposition a representation as a logical expression or a semantic network.\newline Finally, \textbf{the speech analysis} establishes the links between the different sentences, typically identifying the temporal order of statements.
\newline
\subsubsection{Template instantiation phase}
\textbf{NER} consists of identifying in the text all the entities that can fill a role for an event.
\newline
\textbf{The co-referencing resolution}  is mainly based on the identification of similar references despite possible variations and on the resolution of anonymous pronouns to identify the matches between entity references and pronouns referring to them. In particular, it's possible to avoid redundancies and to disambiguate pronouns for extracting events and relationships.
\newline
\textbf{Event detection} aims to classify sentences according to a predefined event type. The event classification is then generally assimilated to the detection of triggers within the sentence.
Once the sentence associated with
an event type given via the extraction of a trigger,
it remains to identify the entities playing a role in it.
This task consists in predicting, for each NE,
its role in the event.
\begin{example}\textbf{Template filling}
\begin{figure}[htt]
  \begin{center}
    \mbox{\includegraphics[width=18cm, height=6cm]{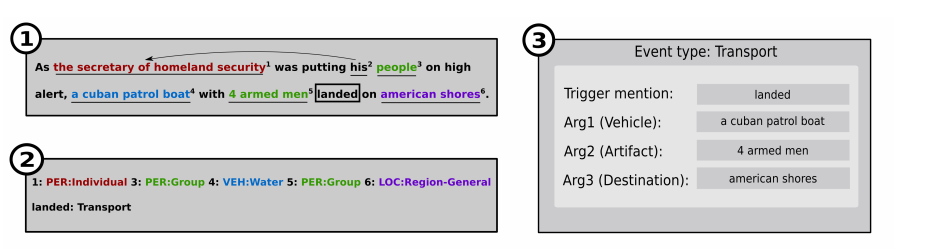}}
      \caption{Template filling example}
    \label{Template filling system example}
  \end{center}
\end{figure}
\end{example}
\newline
NER identifies the different entity references (underlined in box (1)) of the sentence and their type (box (2)). The resolution of co-references makes it possible to identify (here by an arrow) when a pronoun or a reference refers to the same entity. Event detection identifies triggers in the sentence and associates a type with them. The trigger type indicates the type of the template form (3). The arguments in this form are then selected from the previously identified entities.
\newpage
\section{knowledge capitalization system}
This approach \cite{maReference45} represents an automatic event extraction model based on the combination of several current approaches to IE. It is based on the extraction of a set of events of interest from Press news in English and French.
\subsection{Architecture}
\begin{figure}[htt]
  \begin{center}
    \mbox{\includegraphics[width=18cm, height=9cm]{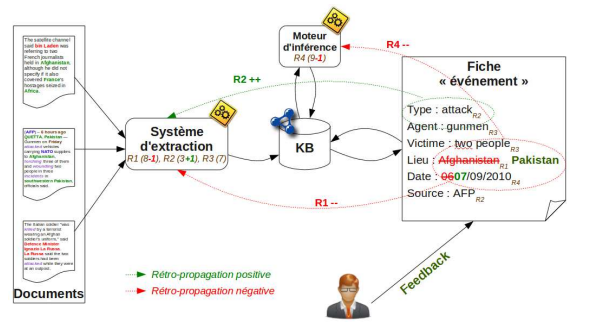}}
      \caption{Architecture of knowledge capitalization system}
    \label{Architecture of knowledge capitalization system}
  \end{center}
\end{figure}

Figure \ref{Architecture of knowledge capitalization system} provides an overview of an overall operation of the knowledge capitalization system.\newline To summarize, a set of documents is processed by the IE system, and the extracted information is stored as RDF triplets in a knowledge base. This is governed by an ontology  and coupled to an inference engine that allows the discovery of a new knowledge. Each event is then presented to the user in a form sheet that he can be  modified according to his own knowledge. 
\newpage
\subsection{Principle}

The knowledge capitalization system proposes the creation of an ontology of the domain that has been developed to model an event and specify its class through different subclasses.\newline
In this approach an event is considered as a triplet E <S, I, SP> \begin{itemize}
\item S is the semantic property and corresponds to the different types of events defined by the domain ontology. 
\item I the temporal component which constitutes the date or the period of occurrence.
\item SP is the spatial property.
\end{itemize}

In a first step, this approach consists of treating an event through a set of sub-events knowing that a sub-event is characterized by the association between "event name" and one or more interesting entities (date, place )\newline
Then, after finding this set, the goal is to automatically merge those which refers in reality to one and the same event.\newline
In a second step, in order to realize the extraction mechanism, this system proposes a process based on the combination of three extractors:\newline
\textbf{A symbolic extractor}  which consists of a set of grammar rules written manually.\newline 
\textbf{A statistical extractor}  in which a model is applied in order to carry out the learning process.
the latter is used taking into account three characteristics:\newline
\begin{itemize}
\item  The inflected form of a word.
\item  The grammatical category.
\item  The type of entity to which it refers.
\end{itemize}
\textbf{An hybrid extractor} which matches symbolic and statistical methods to automatically learn the patterns.\newline
Finally after extracting a set of information, this information will be stored in a knowledge base. This one is governed by an ontology and coupled to an inference engine allowing the discovery of new knowledge.
\newline
The evaluation of the extractions quality is done by a set of rules defined in the ontology and combined with a probabilistic database.\newline
First of all, the form has several fields:\begin{itemize}
\item A general description (type, name, alias),
\item The situation in which this event occurred (place, date/ period), 
\item The various participants involved in the event 
\item Some number of links to other events discovered through the inference engine.
\end{itemize}
The user has the possibility to modify and/or validate these different fields. He can then perform several types of action:
\begin{itemize}
\item Validate a field or the complete form.
\item Add a new field.
\item Correct or delete a field.
\end{itemize}
\newpage
\section{Two-level approach for extracting events} 
The main interest of this approach \cite{maReference43} is to make a correspondence between two levels, a first level's learning (POSITION, PERSON, ORG ) as well as a second level's learning (NEW POSITION, COMING PERSON, IN\_ORG) by using a CRF tool and a double generation of the classifier.
\subsection{Architecture} 
\begin{figure}[htt]
  \begin{center}
    \mbox{\includegraphics[width=18cm, height=10cm]{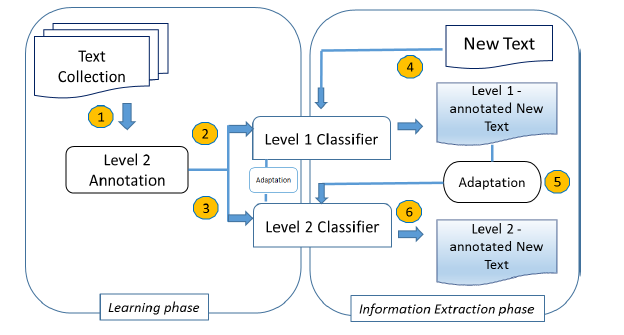}}
      \caption{Two-level approach architecture}
    \label{Two-level approach architecture}
  \end{center}
\end{figure}

The adaptive model considers the level's importance in  IE task. It applies
the same classifier for different levels  by considering an adaptation phase as
depicted in Figure \ref{Two-level approach architecture}.
\newpage
\subsection{Principle}
\begin{figure}[htt]
  \begin{center}
    \mbox{\includegraphics[width=18cm, height=4cm]{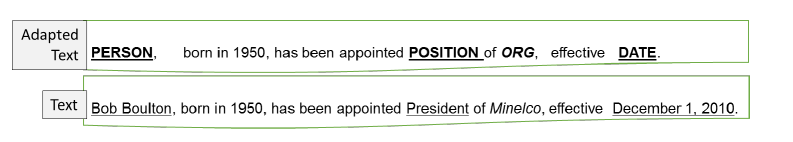}}
      \caption{Adaptation example}
    \label{Adaptation example}
  \end{center}
\end{figure}

\begin{figure}[htt]
  \begin{center}
    \mbox{\includegraphics[width=18cm, height=10cm]{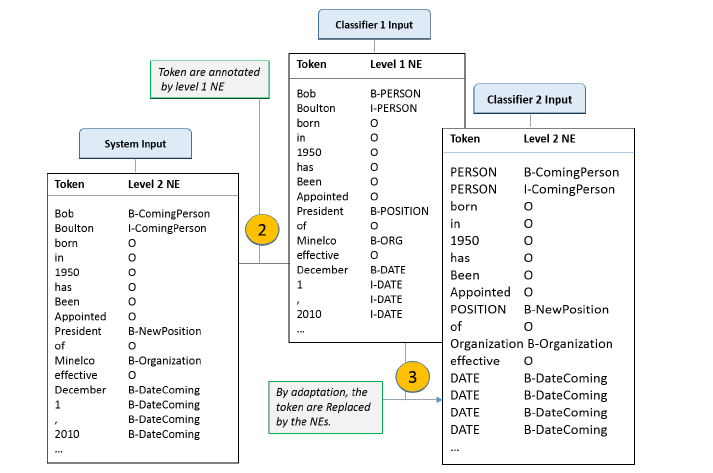}}
      \caption{Inputs preparation}
    \label{Inputs preparation}
  \end{center}
\end{figure}

\begin{figure}[htt]
  \begin{center}
    \mbox{\includegraphics[width=18cm, height=5cm]{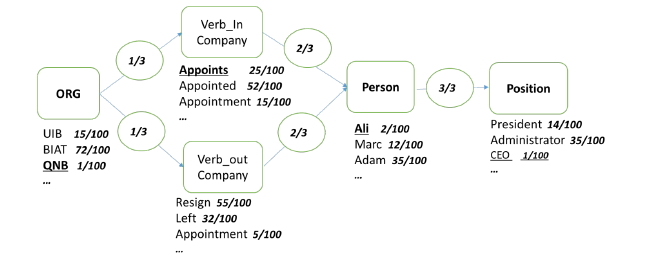}}
      \caption{Adaptive model for level1}
    \label{Adaptive model for level1}
  \end{center}
\end{figure}

\begin{figure}[htt]
  \begin{center}
    \mbox{\includegraphics[width=18cm, height=5cm]{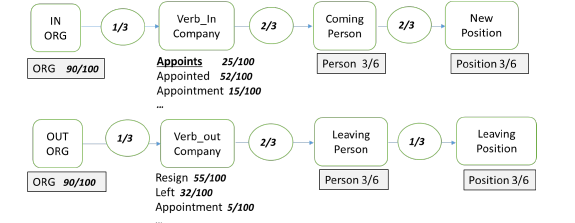}}
      \caption{Adaptive model for level2}
    \label{Adaptive model for level2}
  \end{center}
\end{figure}

The idea of this approach starts from the level 2 annotated corpus  by processing
the level 1 entities. Hence, the learning phase was applied to generate the level 1
classifier. The corpus is also prepared for deriving the level 2 classifier after an adaptive phase. It means replacing all annotated texts by their
corresponding NE. For instance,  as depicted in Figure \ref{Adaptation example}, it shows how
some annotated parts of the text were replaced by their corresponding named entities.
Hence, the level 2 classifier input is prepared for obtaining the final event recognition. Figure \ref{Inputs preparation}  presents
how it was prepared for both classifiers 1 and 2.
\newline
\newpage
 A token labeled with B-T is
the beginning of a named entity of type T while a token labeled with I-T is
inside (but not the beginning of) a named entity of type T. In addition, there
is a label O for tokens outside of any named entity. From the system input, the
classifier 1 input considers only the level 1 NE. By adaptation 
the tokens transformed by their corresponding level 1 NE, thus the level 2 classifier input is prepared.
\newpage
Figures \ref{Adaptive model for level1} and \ref{Adaptive model for level2} present the adaptive models for level
1 and level 2 respectively. In the graph it presents the different tokens,  the transitions
and their corresponding probabilities degrees.
Applying the CRF-based classifier, means finding the annotation y and
maximizing the probability p(y|x).
\newline
Let's consider the following text: x="QNB appoints Ali as CEO".
\newline
\newline
\textbf{For level1:}\newline
We can obtain the following annotations:\newline
Y$^{'l1}_1$ ="ORG Verb-IN-Company Person Position"\newline
Y$^{'l1}_2$ ="ORG Verb-OUT-Company Person Position"\newline
x'="ORG Appoints Person Position" as a new input for level 2 classifier.
\newpage
\textbf{For level2:}\newline
We can obtain the following annotations:
\newline
Y$^{'l2}_1$ ="InORG Verb-IN-Company ComingPerson NewPosition"
\newline
Y$^{'l2}_2$ ="OutORG VerbOutCompany LeavingPerson LeavingPosition"
\newline
Les probabilités qui sont obtenues:
\begin{figure}[htt]
  \begin{center}
    \mbox{\includegraphics[width=15cm, height=4cm]{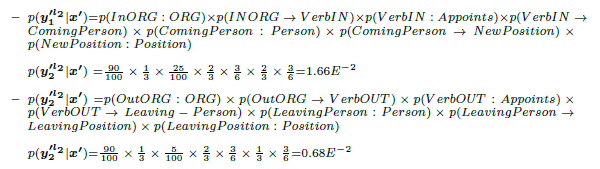}}
 \end{center}
\end{figure}
\section{Comparative analysis}
We present below a recapitulatory table that shows all the work on event extraction from the approaches we studied.
This study was carried out according to the methods used for event extraction, the language applied as well as the advantages and disadvantages of each of the approaches.

\newpage
\begin{table}[!ht]
\centering
\begin{tabular}{|p{3cm}|p{2cm}|p{2cm}|p{2.5cm}|p{2.5cm}|}
\hline
Approach& Language& Method &Advantage&Disadvantage \\
\hline
GLAEE &English &Knowledge-driven approach&The average recognition rate is 68.93\% 
&The keywords are defined by the expert\\
\hline
CRF & All the languages& Data-driven approach &Gives a probability to each returned extraction is an index of quality of the information&Difficult to access and to be modified\\
\hline
Template filling through IE & French&Knowledge-driven approach&Identifying multiple events within a sentence
&The rules are written manually \\
\hline
Knowledge capitalization system & English&Hybrid approach&The use of an inference engine allowing the discovery of new knowledge
&The rules are written manually \\
\hline
Two-level approach for extracting events & English &Data-driven approach&The average recognition rate is 67.91\%&Depends on the quantity and the quality of the learning data\\
\hline
\end{tabular}
\caption{Recapitulatory table}
\label{table:5}
\end{table}

\section{Conclusion}
A review of the scientific literature was conducted, based on several reference documents in the field. This study allowed us to develop the most useful theoretical concepts for our research work.
The event extraction approaches depend on the additional learning data provided by the domain expert.
CRFs are among the approaches that have proven their efficiencies in the field of Level 1 and Level 2 of IE.
In the next chapter we propose a new approach that allows the use of OIE system and ontology.
\chapter{New approach: Event extraction based on OIE and ontology}
\section{Introduction}
In the previous chapter, we presented some existing work in the event extraction field. As part of this work we will present our approach, then we will describe the algorithms used for the processing of the extracted information.
\newline
Event extraction approaches need a domain knowledge to generate rules or patterns of recognition. In particular, in the GLAEE approach, the user should define the list of keywords and cue words to produce a set of patterns.
\newline
The CRF-based event extraction approach has proved its effectiveness in NER (level1) and event extraction (level2). Among the advantages of CRFs is to use a statistic model that does not require the additional knowledge provided by a domain expert.
\newline
Our goal in this work is to reduce human intervention during  event extraction. In particular we present our approach which is based on two phases: A learning phase and a recognition phase.
\section{System architecture}
The main interest of this approach is how to extract a specific information from all existing relationships between all entities that can be found in an open corpus.
\newline
Our approach admits 2 phases that depend on each other. The first phase is the modeling of an event by an ontology and  constructing a set of  rules acquired manually .
The second is the recognition phase which includes the RE, the NER and an automatic
reasoning between learning rules and an input ontology adaptation. Our aim from two these phases is an eventually event extraction.
\begin{figure}[htt]
  \begin{center}
    \mbox{\includegraphics[width=15cm, height=10cm]{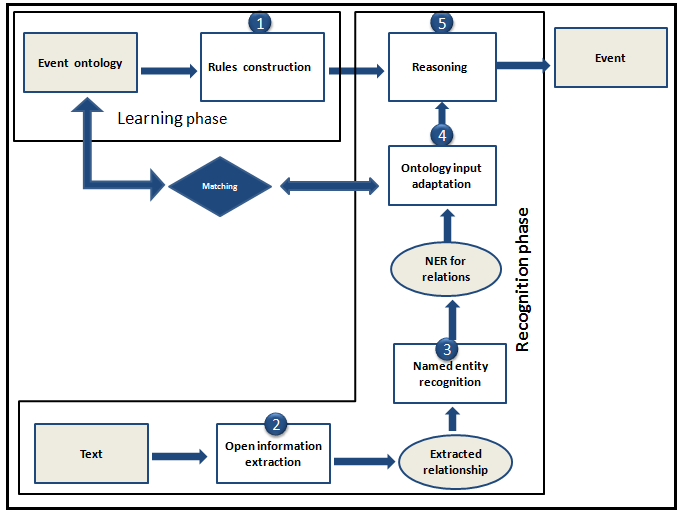}}
      \caption{System architecture}
    \label{System architecture}
  \end{center}
\end{figure}
\subsection{Learning phase}
The notion of an event does not admit a strict and precise definition, we always find the following notions:  Movement over time, modification of a situation, etc. From here we can consider that an event is an object  admits an existence in the space of time and  depends on other objects in relation, that's why we find few approaches in which events are modeled by an ontology \cite{maReference45} and this is our case .
\newline
An ontology is a set of concepts in a domain, as well as relationships between these concepts.
\newline
In our case  the concepts represent the named   entities of the event (Person, Organization...) and their roles (Coming\_person, Leaving\_person, IN\_ORG) being in certain relationships.
\newline 
The rules construction is an important step in our approach which allows, through the ontology  and the result of the recognition phase, a possible event extraction.
\subsection{Recognition phase}
The recognition phase combines four steps which are as follows: OIE for a relation extraction, NER, ontology input adaptation and reasoning.
\subsubsection{Open information extraction}
The input of the system is a text in natural language, the first step of recognition is generated by an OIE system which allows an extracting textual relationship triplets present in each sentence. The relationship triplets contain three textual components (Arg1, Rel, Arg2) where the first and the third indicate the pair of  arguments and the second indicates the relationship between them.
The aim of this step is to restrict the content of the text into relationships that are well defined and have a specific meaning for each sentence in the text.
\subsubsection{Named entity recognition}
The input of the NER tool is a triplet found in the previous step.
The triplet (Arg1, Rel, Arg2) will have an automatic recognition of named entities after a tokenization step.
In this step, the system can detect person, organization, location, etc. in any part of the triplet.
\subsubsection{Ontology input adaptation}
As we saw in the previous chapter there are two types of OIE systems, the first extracts only triplets      
 \textbf{(Arg1; verb; Arg2)}, the second handles triplets with attribution and  conditions if they exist. 
It is in the form of \textbf{(Arg1; verb; Arg2) [attribution/condition].} \newline
After recognizing the NEs,  the verbs will be passed through a lemmatization layer that will convert conjugated verbs to their infinitive form.
Every token recognized by a NE will be added as instance in the ontology.\newline
In the learning phase we said that the concepts represent the NEs of the event and their roles.\newline
The principle of our approch in this step that the tokens will be added as instances to the ontology and linked by relations if the following  conditions are validated:\newline
\begin{itemize}
\item The number of named entities is greater than or equal to 2.
The NE can be found in the triplet or in  the attribution/condition part (if we are in the second type of OIE system)\newline
\item The lemmatized verb and the other relations between delimiters ";" are included in the list of  the ontology relations and NEs can be linked with these relations. 
\end{itemize}
\subsubsection{Reasoning}
The reasoning is a stage after entering the instances and linking them by their specific relations.
\newline The reasoner is a  software which  infers logical consequences from a set of rules to affect for each instance its role (event). 
\section{Extraction process}
In this part, we begin the presentation of three algorithms of our event extraction approach. The \textbf{Algorithm 1} describes the mechanism of learning on a list of named entities, relationships and roles of an event to produce as an output, an ontology and a set of rules.
The \textbf{Algorithm 2} presents the recognition phase.
The \textbf{Algorithm 3} illustrates the adaptation step for the  OIE system and the \textbf{Algorithm 4} is for the matching step.
\subsection{Learning phase}
The learning algorithm is based on building an ontology
by a list of named entities that represent the classes, the roles represent  the subclasses and a set of relationships.
using in parallel these classes, subclasses and relationships to build the rules.\newline
\newline
\algblock{Input}{EndInput}
\algnotext{EndInput}
\algblock{Output}{EndOutput}
\algnotext{EndOutput}
\newcommand{\Desc}[2]{\State \makebox[2em][l]{#1}#2}
\begin{algorithm}\label{alg:Learning phase}
\caption{Learning phase}
\begin{algorithmic}
  \Input
  \Desc{L1 :}{List of entities}
  \Desc{L2 :}{List of roles}
  \Desc{R :}{List of relations}
  \EndInput
  \Output
  \Desc{O :}{ Ontology}
  \Desc{RC :}{ Rules Construction}
  \EndOutput
  \ForEach {$ \text{Entity i in L1}$ \textbf{do}}
\State\,\,\,\quad$Class \gets Class\cup add(i)$  \textit{//Add named entities as classes in the ontology}
\ForEachesp {$ \text{Role j in L2}$ \textbf{do}}
\State \quad\quad\quad$Subclass \gets Subclass \cup add(i,Class)$  \textit{// Add roles as subclasses for named entities that represent classes in ontology}
\Endesp
\EndForEach
 \ForEach {$ \text{Relation r in R}$ \textbf{do}}
\State $Relation \gets Relation\cup add(r)$ \textit{//Add a set of relationships to the ontology}
\EndForEach  
\State $O \gets Event\_Ontology(Class,Subclass,Relation)$  \textit{//The event is modeled by these classes, subclasses and relationships}
\State $RC \gets  
Rules\_Construction(Class,Subclass,Relation)$ \textit{//The rules are made by these classes, subclasses and by relationships}
\end{algorithmic}
\end{algorithm}
\newpage
\subsection{Recognition phase}
The recognition algorithm admits as input a text for the OIE system. It applies as a first step the  extraction of relation triplets. After a tokenization layer of triplets, every token will admit a NER step.
Finally the result of the adaptation is the entry of the reasoning with the set of rules. The final result of the reasoning is a set of events.
\newline
\newline
\begin{algorithm}\label{algo:Recognition phase}
\caption{Recognition phase}
\begin{algorithmic}
  \Input
  \Desc{T :}{Text}
  \Desc{RC :}{Rules Construction}
  \Desc{RT :}{Relation Triplet}
  \EndInput
  \Output
  \Desc{E :}{Events}
  \EndOutput
  \State $\textbf{Begin}$
\State \quad\quad$RT \gets OIE(T)$ \textit{//Relation extraction by an open information extraction tool} 
\State \quad\quad$R \gets RT.substring(pos(";",RT)+1,-pos(";",RT)-1)$ \textit{//Extract the verbal part of the relationships}
 \ForEach{$ \text{Relation\_triplet rt in RT}$ \textbf{do}}
\State \quad\quad$Token \gets Tokenization(rt)$ \textit{//Cut the relationship triplet into tokens}
\ForEachesp {$ \text{Token t in rt}$ \textbf{do}}
\State \quad\quad\quad$NE \gets NER(t)$ \textit{//Named entity recognition automatically}
\State \quad\quad\quad$A \gets Adaptation(NE,R,O)$
\State \quad\quad\quad$RS \gets Reasoning(A,RC)$
\State \quad\quad\quad$E \gets Event\_Extraction(RS)$ \textit{// Event extraction By a reasoner}
\IFend
\EndForEach
\end{algorithmic}
\end{algorithm}
\newpage
The adaptation algorithm is used to take  the token of every recognized named entity as input to the ontology if the number of named entities is greater than or equal to 2 and if the relation is included in the relationship list of the ontology.
\newline
\newline
\begin{algorithm}\label{algo:Adaptation }
\caption{Adaptation }
\begin{algorithmic}
  \Input
  \Desc{NE :}{ Named entity }
  \Desc{R :}{Relation}
  \Desc{O :}{Event\_Ontology}
  \EndInput
  \Output
  \Desc{tk :}{tokens\_input\_ontology }
  \Desc{RL :}{Relation\_lemmatizated }
  \EndOutput
  \State $\textbf{Begin}$\State  
\IF {$ \text{(triplet.Count(NE)>=2)}$  \textbf{then}}
\State \quad\quad\quad\quad$tk \gets NE.token$ \textit{// tk take tokens that are recognized by named entities}

\IFee {$ \text{(R in O.Relations)}$  \textbf{then}}
\State  \quad\quad\quad\quad\quad$RL \gets R.Lemmatization$ \textit{// Verbs are transformed into infinitive}
\State \quad\quad\quad\quad\quad$ Matching(tk,RL,O)$
\IFeeend
\IFend
\IFfend
\end{algorithmic}
\end{algorithm}
\newpage
The matching algorithm is used to affect tokens as instances in the ontology and linked them by relations.
\newline
\newline
\begin{algorithm}\label{algo:Matching }
\caption{Matching}
\begin{algorithmic}
  \Input
  \Desc{O :}{Ontology}
  \Desc{tk :}{Tokens\_input\_ontology}
  \Desc{R :}{Relation}
  \EndInput
  \Output
  \Desc{O2 :}{Ontology with instances linked by relations}
  \EndOutput
  \State {$\textbf{Begin}$}
 \IFf {$ \text{(O.Class==tk.NE)}$  \textbf{then}} \textit{// if the named entity of the token matches a class of the ontology}
\State \quad\quad$tk.addAsInstanceOf(O.Class)$ \textit{// the token will be added as instance under the class }
\State \quad\quad$tk.LinkInstancesBy(R)$\textit{// the tokens will be linked by a relation R }
\IFfend
\State \quad\quad$O2\gets New\_OntologyWith(tk,R)$\textit{// a new ontology with instances tk linked by relation R}
\IFfend
\end{algorithmic}
\end{algorithm}
\section{Illustrative examples}
This part presents an example that describes the different steps of each algorithm.
\subsection{Learning phase}
\begin{figure}[htt]
  \begin{center}
    \mbox{\includegraphics[width=14cm, height=7cm]{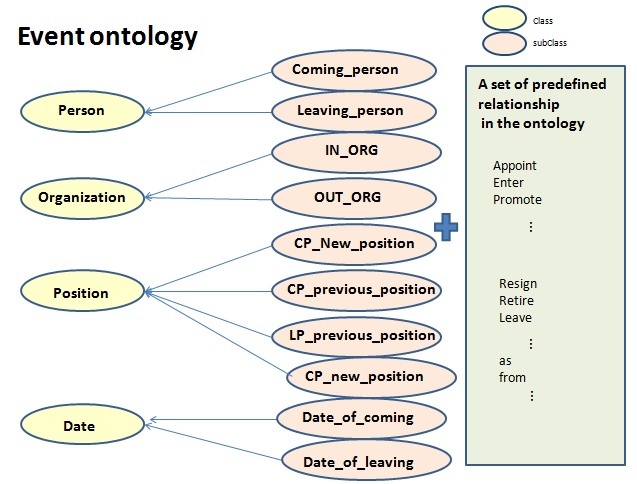}}
      \caption{Example of an Event ontology }
    \label{Example of an Event ontology}
  \end{center}
\end{figure}

\begin{figure}[htt]
  \begin{center}
    \mbox{\includegraphics[width=14cm, height=3cm]{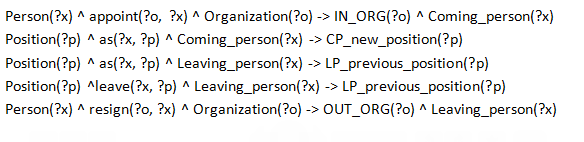}}
      \caption{Example of rules construction }
    \label{Example of rules construction }
  \end{center}
\end{figure}

Figures \ref{Example of an Event ontology} and \ref{Example of rules construction } illustrate successively the learning phase which composed of event ontology (classes, subclasses and relations) and rules construction.
\newline
The class Person has two subclasses: Coming\_person and Leaving\_person.
\newline
The class Organization has two subclasses: IN\_ORG and OUT\_ORG.
\newline
The class Position has four subclasses : CP\_new\_position , CP\_previous\_position , LP\_previous\_position, LP\_new\_position.
\newline
The class Date has two subclasses : Date\_of\_coming, Date of leaving.
\newline
Our ontology contains a set of predefined relationships to connect two eventual instances. 
\newline
A set of rules are predefined to affect for every instance its role.\newline
For example: \newline
\textbf{Person(?x) $\wedge$ appoint(?o,?x) $\wedge$ Organization(?o)$\rightarrow$ IN\_ORG(?o)  $\wedge$ Coming\_Person(?x)} means any instance x of type Person is connected by an 'appoint' relation with any instance of type Organization o gives us the result that the  instance x is a Coming\_person and the instance o is an  IN\_ORG.
\newline
\newpage
\subsection{Recognition phase}
Each text will be applied to an OIE system, followed by an  automatic recognition of named entities which include a tokenization step,  taking into account the stage of ontology input adaptation which include a lemmatization step.
\newline
\subsubsection{First type of an OIE system} 
\begin{figure}[htt]
  \begin{center}
  \mbox{\includegraphics[width=14cm, height=6.5cm]{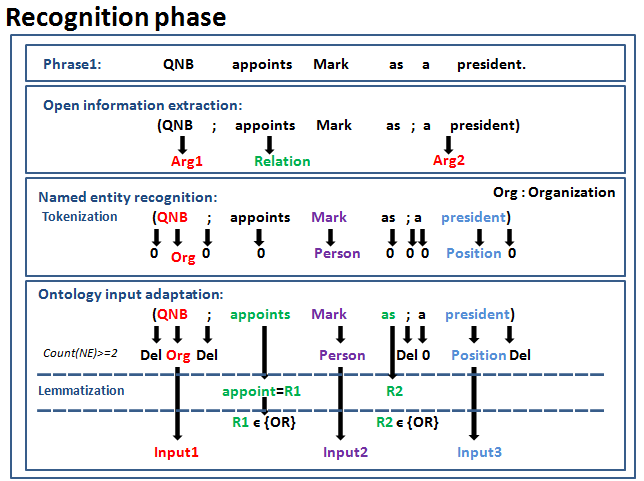}}
      \caption{Example of recognition phase for the phrase " QNB appoints Mark as a president " }
    \label{Example of recognition phase for the phrase " QNB appoints Mark as a president "}
  \end{center}
\end{figure}

\begin{figure}[htt]
  \begin{center}    \mbox{\includegraphics[width=14cm, height=6cm]{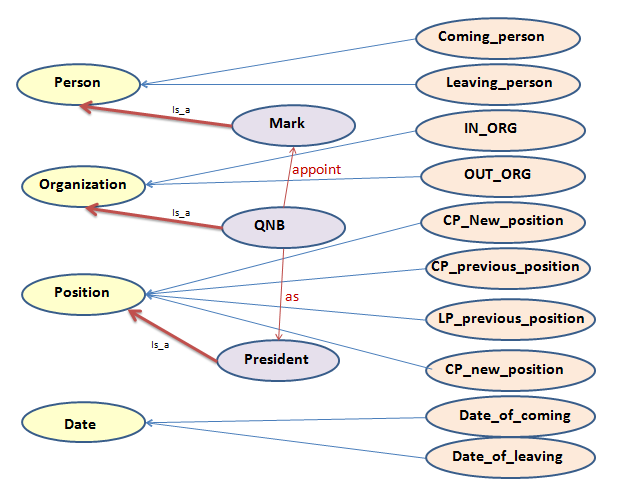}}
      \caption{Example of ontology after adaptation step for the phrase1 }
    \label{Example of ontology after adaptation step for the phrase1}
  \end{center}
\end{figure}

\begin{figure}[htt]
  \begin{center}
    \mbox{\includegraphics[width=14cm, height=8cm]{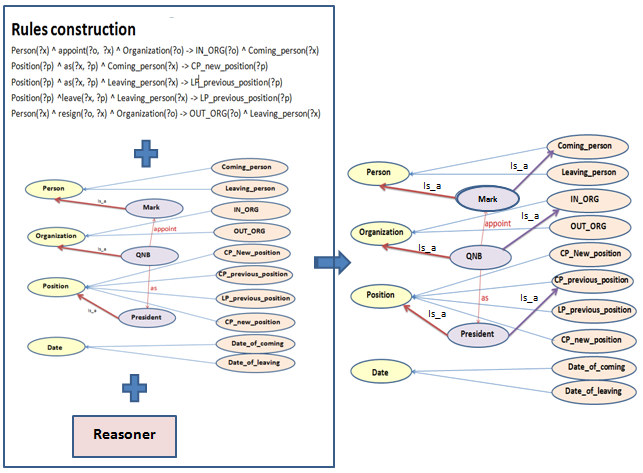}}
      \caption{Example of reasoning step and event extraction for the phrase1}
    \label{Example of reasoning step and event extraction for the phrase1}
  \end{center}
\end{figure}

\begin{figure}[htt]
\begin{center}
 \mbox{\includegraphics[width=14cm, height=6.5cm]{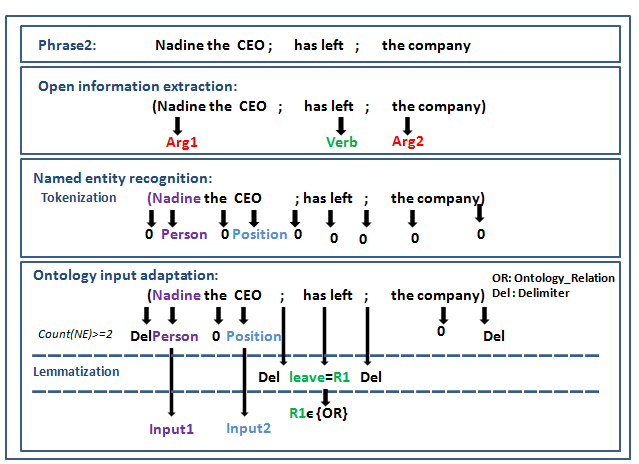}}
      \caption{Example of recognition phase for the phrase " Nadine the CEO has left the company" }
    \label{Example of recognition phase for the phrase " Nadine the CEO has left the company"}
  \end{center}
\end{figure}

\begin{figure}[htt]
  \begin{center}
    \mbox{\includegraphics[width=14cm, height=6cm]{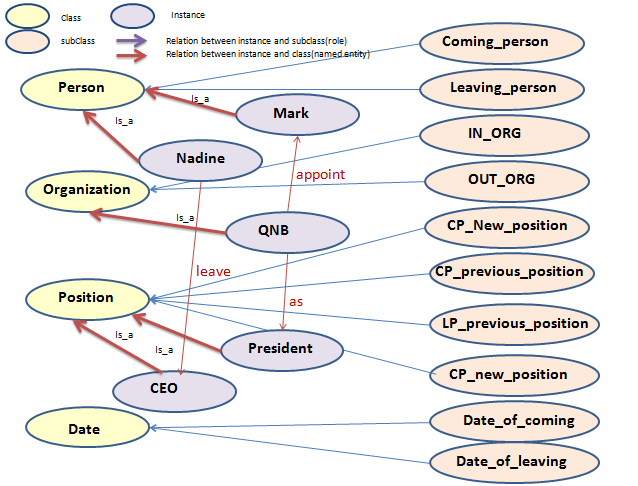}}
      \caption{Example of ontology after adaptation step for the phrase2 }
    \label{Example of ontology after adaptation step for the phrase2}
  \end{center}
\end{figure}

\textbf{Phrase1:} 
After recognizing QNB as an Organization, Mark as a person and president as a position.\newline
The first condition is checked:  The number of named entities is greater than or equal to 2.
The part between the delimiters ";" presents the part that contains the relations.
\newline
In this example the verb "appoints" after being lemmatized and the relation "as" are compared  to the list of relations in the ontology.\newline \newline
The second condition is verified: "appoint" and "as" are in the list of ontology relations as shown in Figure \ref{Example of recognition phase for the phrase " QNB appoints Mark as a president "}.\newline \newline
The tokens that are recognized by named entities (QNB, Mark and president)  will be entered as instances under the classes of the ontology and  will be connected by their corresponding relationships (appoint, as) as shown in Figure \ref{Example of ontology after adaptation step for the phrase1}.
\newline \newline
After reasoning, the entries (input1 , input2, input3) are automatically linked by their roles (event)
in this example QNB has the role of IN\_ORG, 
Mark has the role of Coming person, 
president has the role of CP\_new\_position as shown in Figure \ref{Example of reasoning step and event extraction for the phrase1}.
\newpage
\newpage
\textbf{Phrase2:} In this example Nadine is recognized  as a person and CEO as a position.
The first condition is checked: The number of named entities is greater than or equal to 2.
The part between the delimiters ";" presents the part that contains the relations.
\newline
In this example the verb "has left" after being lemmatized "leave" is compared  to the list of relations in the ontology.\newline 
\newline
The second condition is verified: "leave" is in the list of the ontology relations as shown in Figure \ref{Example of recognition phase for the phrase " Nadine the CEO has left the company"}.\newline \newline
Tokens which are recognized by named entities (Nadine, CEO)  will be entered as instances under the classes (Person, Position) of the ontology and  will be connected by their corresponding relationships (leave) as shown in Figure \ref{Example of ontology after adaptation step for the phrase2}.
\newpage

\begin{figure}[htt]
  \begin{center}
    \mbox{\includegraphics[width=15cm, height=10cm]{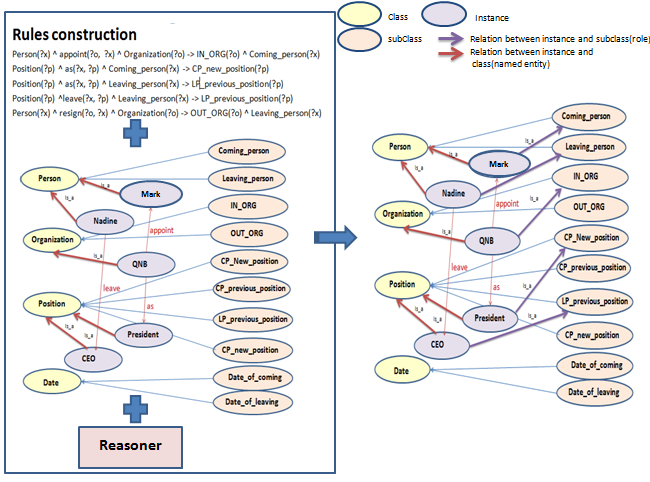}}
      \caption{Example of reasoning step and event extraction for the phrase2}
    \label{Example of reasoning step and event extraction for the phrase2}
  \end{center}
\end{figure}

\newpage
After reasoning, the entries (input1, input2, input3) are automatically linked by their roles (event)
in this example Nadine has the role of Leaving\_person and CEO has the role 
of LP\_previous\_position as shown in figure \ref{Example of reasoning step and event extraction for the phrase2}.
\newpage
\begin{figure}[htt]
\subsubsection{Second type of an OIE system}  
  \begin{center}
\mbox{\includegraphics[width=14cm, height=10cm]{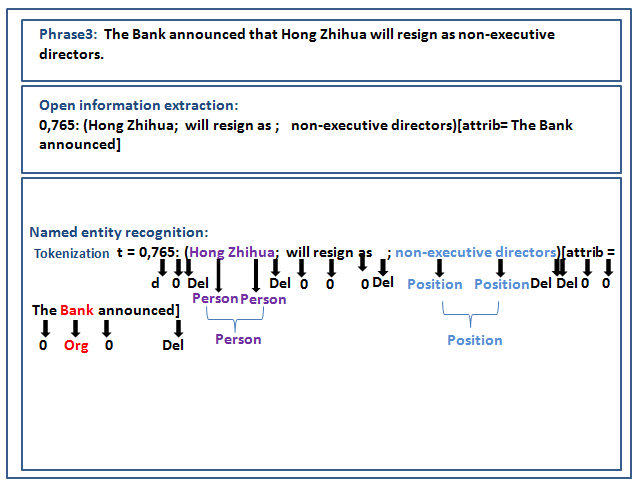}}
      \caption{Recognition phase for the second type of OIE system}
    \label{Recognition phase for the second type of OIE system}
  \end{center}
\end{figure}
In this example the number of entities is greater than 2 and can be linked by a relationship.
So input1, input2 and input3 should be entered to the ontology as instances under the classes Person, Position and Organization.
if there are triplets that share the same verbal part, we work on the triplet which has the highest degree of certainty \textbf{d}.
\newpage
\begin{figure}[ht]
  \begin{center}
    \mbox{\includegraphics[width=14cm, height=8cm]{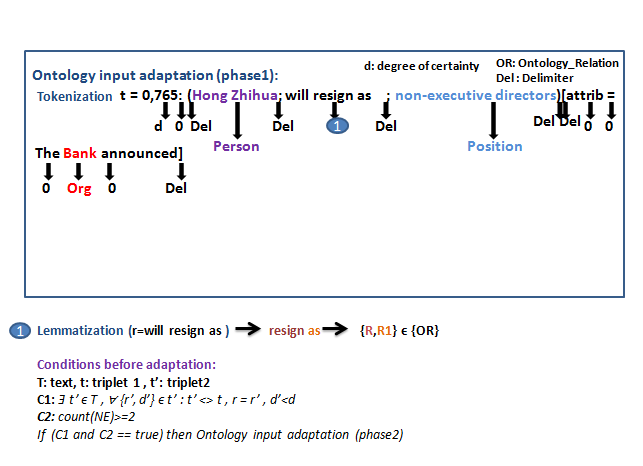}}
      \caption{Ontology input adaptation for the second type of OIE system (phase1) }
    \label{Ontology input adaptation for the second type of OIE system (phase1)}
  \end{center}
\end{figure}
\begin{figure}[ht]
  \begin{center}
    \mbox{\includegraphics[width=14cm, height=4.5cm]{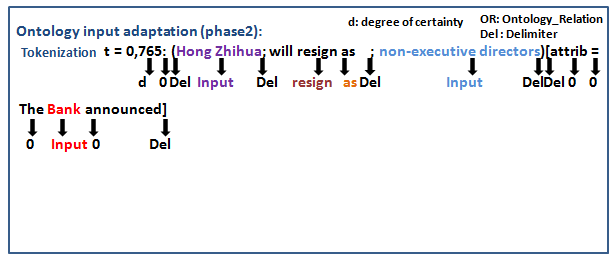}}
      \caption{Ontology input adaptation for the second type of OIE system (phase2) }
    \label{Ontology input adaptation for the second type of OIE system (phase2)}
  \end{center}
\end{figure}
\newpage
\begin{figure}[ht]
  \begin{center}
    \mbox{\includegraphics[width=14cm, height=10cm]{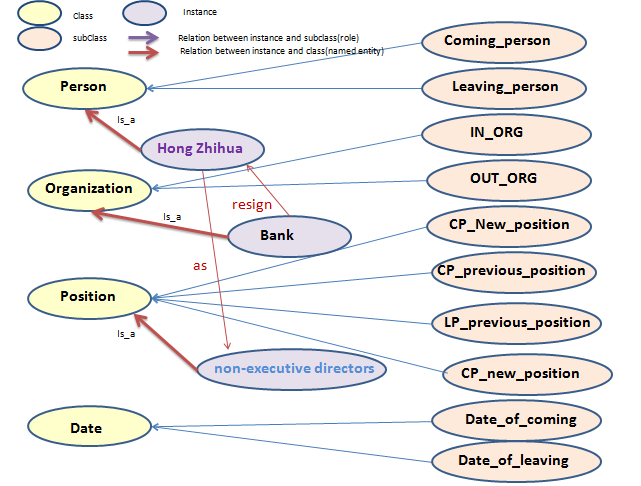}}
      \caption{Example of ontology after adaptation step for the phrase3 }
    \label{Example of ontology after adaptation step for the phrase3}
  \end{center}
\end{figure}
The reasoner reasons through a set of  rules to finally conclude that Hang Zhihua is a Leaving\_person, non-executive directors is a LP\_previous\_position and Bank is an OUT\_ORG.
\newpage
\begin{figure}[ht]
  \begin{center}
\mbox{\includegraphics[width=15cm, height=9cm]{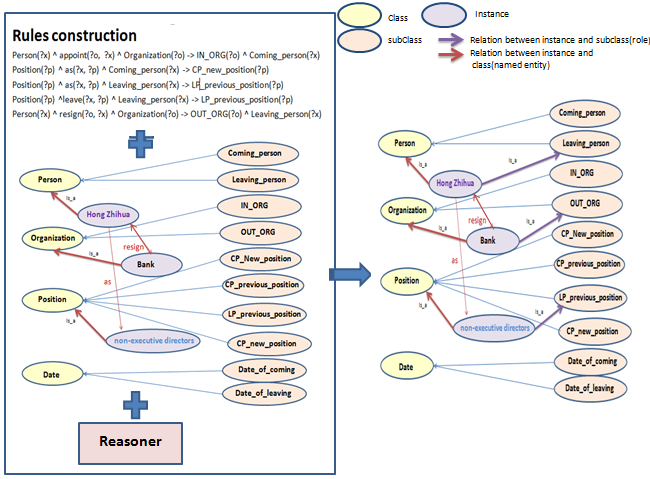}}
      \caption{Event Extraction for the phrase3 }
    \label{Event Extraction for the phrase3}
  \end{center}
\end{figure}
\section{Conclusion}
In this chapter, we defined the motivation for choosing the system architecture, the practical steps which are divided into two main phases: The learning and the recognition phase. \newline
Our approach requires several tools related to the study for event extraction as well as the ontology.
\newline
We have brought our approach as well as the characteristics and the technical details of the elaboration of our system, let's go to the presentation of the experimental study.
\chapter{Experimental study}
\section{Introduction}
This chapter presents the experimental work and the evaluation concerning the relevance of our system detailed in the previous chapter. In fact, we express the automatic extraction of relation triplets by an OIE system as well as the event extraction  by applying them to a set of corpus reserved for tests through an ontology.
\section{Test metrics}
In our evaluation, we chose the metrics that will allow us to evaluate the results of our work and measure the performance of the proposed system.\newline
When the system returns an answer to a text and a class, two choices are available:\newline
\begin{itemize}
\item The message belongs to the class.\newline
\item The message does not belong to the class\newline
\end{itemize}
\newpage
This gives 4 different possible cases:\newline
\begin{itemize}
\item\textbf{True Positive (TP):} The system correctly finds the message as belonging to the class.\newline
\item\textbf{False positive (FP):} The system mistakenly finds the message as belonging to the class.\newline
\item\textbf{True negative (TN):} The system rightly finds the message as not belonging to the class.
\newline
\item\textbf{False negative (FN):} The system mistakenly finds the message as not belonging to the class.
\end{itemize}
Precision is the measure of quality, recall is the measure of quantity and the F-measure is the synthesis of recall and precision. These three measures were chosen for their frequent use in the field.
\newline
\newline
\textbf{Precision:}
Proportion of relevant solutions that are found. It measures the ability of the system to provide all relevant solutions.
\[Precision
  = \dfrac{TP}{TP+FP}
\]
\newline\newline\newline
\textbf{Recall:}
Proportion of solutions found that are relevant. Measures the ability of the system to reject irrelevant solutions.\newline
\[Recall
  = \dfrac{TP}{TP + FN}
\]
\newline
\textbf{F-measure:}
Measures the ability of the system to provide all relevant solutions and to deny others.
\[F-measure
  = \dfrac{2 * PR}{R + P}
\]
\newline\newline
In fact, the evaluation of our approach is divided into two parts, described in more detail in the following sections. At first, the results of a general evaluation of the recognition rate of all the events. Next, the second part relates to the different types of events.
\section{Implementation}
In this part we start with an example of a test done on an extracted english text in order to observe the mechanism of this approach.
This example describes the steps of our approach and the transition from a simple text written in natural language to an automatically extracted set of events.
\subsection{Learning phase}
For the learning phase we chose protégé as an ontology modeling tool.\newline
Protégé’s plug-in architecture can be adapted to build both simple and complex ontology-based applications. Developers can integrate the output of Protégé with rule systems or other problem solvers to construct a wide range of intelligent systems.
Figure 4.1 presents an ontology for the event management change and Figure 4.2 presents a set of relations in the ontology.
\newpage
\begin{figure}[ht]
  \begin{center}
    \mbox{\includegraphics[width=18cm, height=6cm]{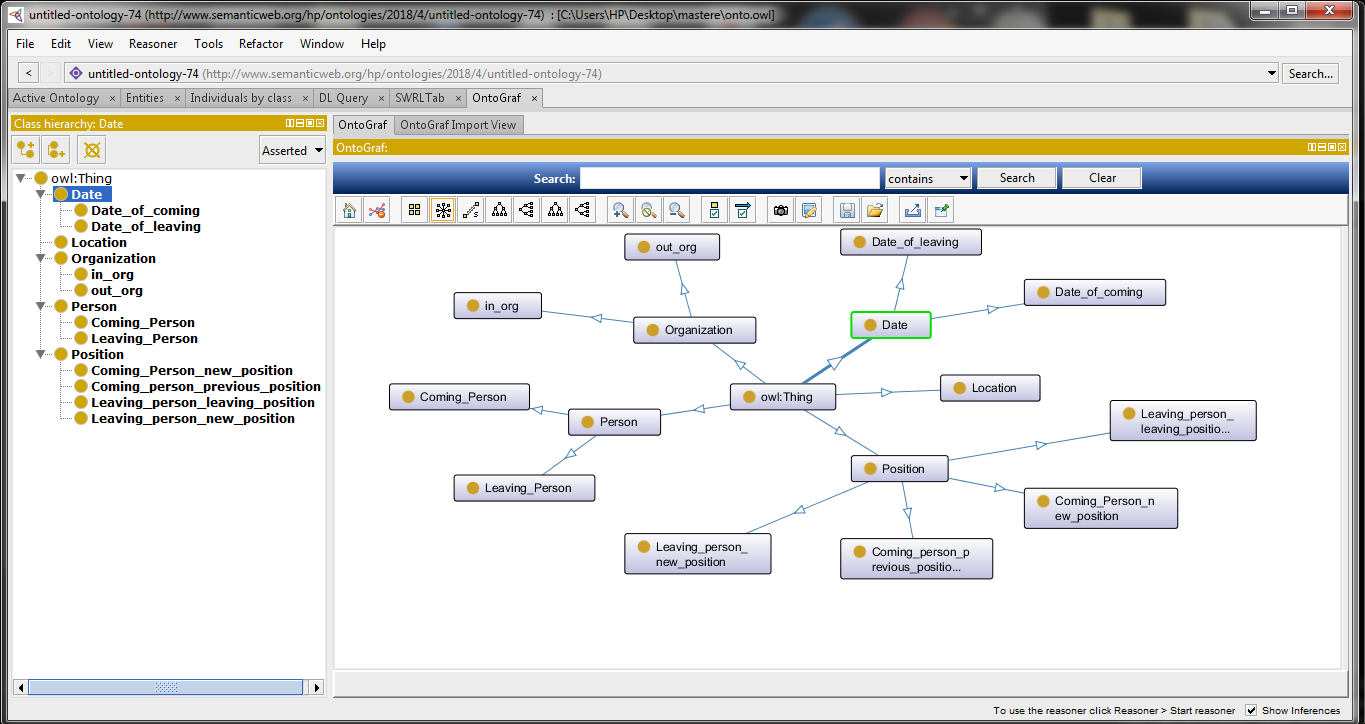}}
      \caption{Management change ontology}
    \label{Management change ontology}
    \end{center}
\end{figure}
\begin{figure}[ht]
  \begin{center}
    \mbox{\includegraphics[width=18cm, height=6.5cm]{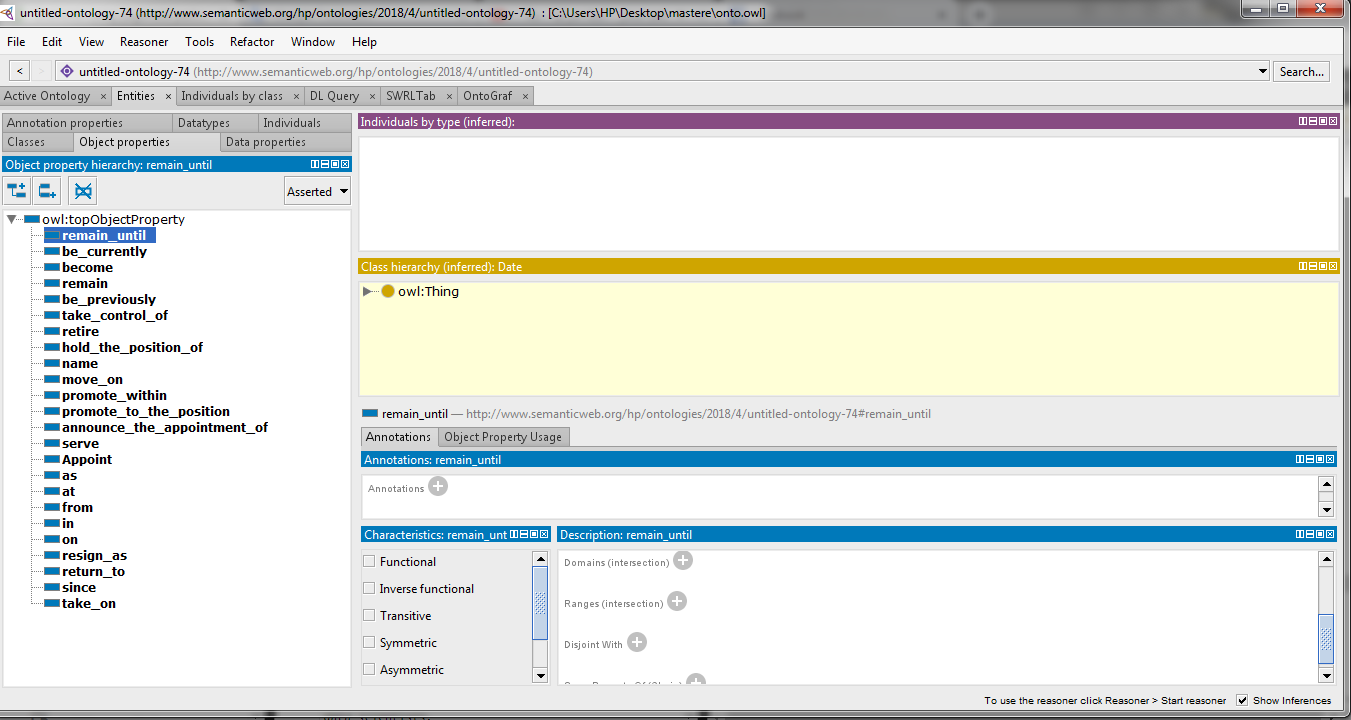}}
      \caption{Relations in the ontology}
    \label{Relations in the ontology}
  \end{center}
\end{figure}
\newpage
For the rules construction we chose SWRLTab in protégé.
The SWRLTab is a Protégé plugin that provides a development environment for working with SWRL rules.\newline
A rule axiom consists of an antecedent (body) and a consequent (head), each rule consists of a (possibly empty) set of atoms. A rule axiom can also be assigned by an URI reference, which could serve to identify it.
\newline
Informally, a rule may be read as meaning that if the antecedent holds (is "true"), then the consequent must also hold.\newline
Figure \ref{Rules construction} shows the rules we used in our application.
\begin{figure}[ht]
  \begin{center}
    \mbox{\includegraphics[width=18cm, height=10cm]{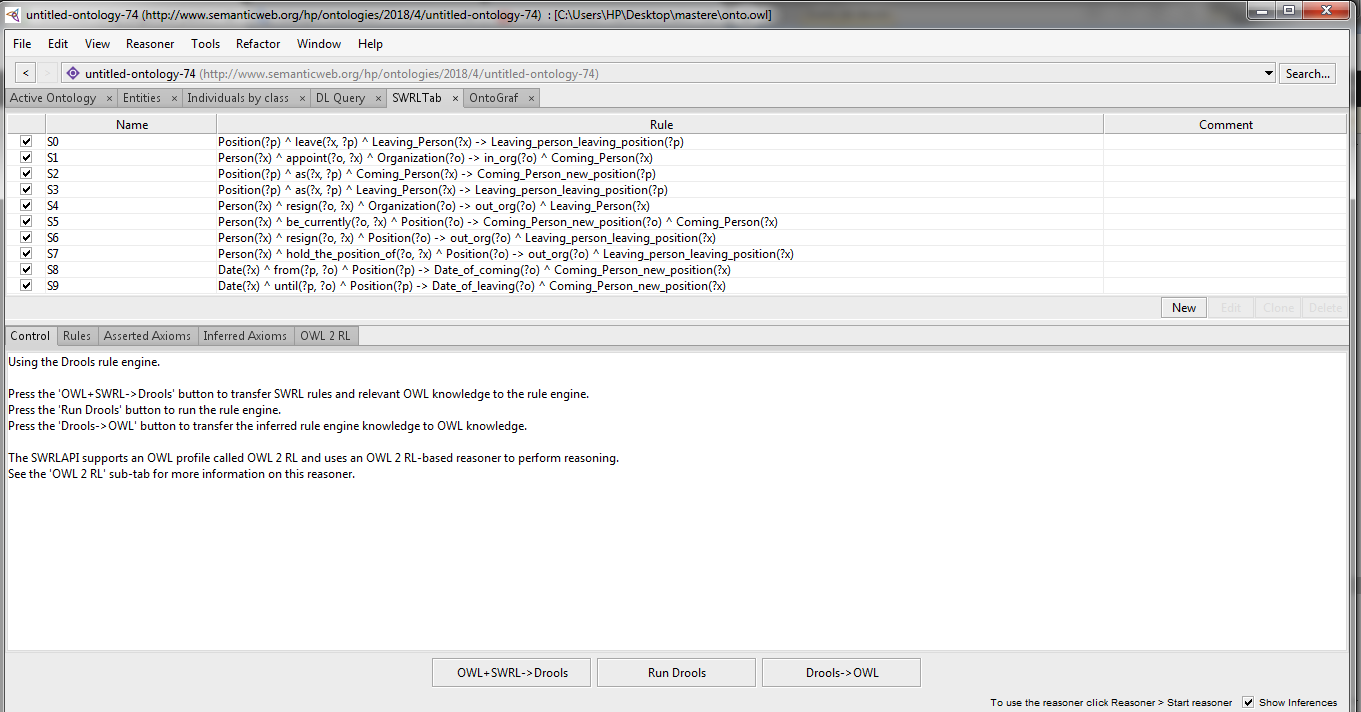}}
      \caption{Rules construction}
    \label{Rules construction}
    \end{center}
\end{figure}
\newpage
\subsection{Recognition phase}
In this part we used OLLIE as an OIE,  we used Python as one of the languages commonly used for Automatic Language Processing, and it is very fashionable right now.
\newline
Its particularity is to be open source and to benefit from an active community of users who collaborate and share their solutions to solve common problems and especially "libraries" which contain different algorithms directly usable by data scientists.
among these libraries we can mention spacy and nltk \footnote{https://www.ekino.com/articles/handson-de-quelques-taches-courantes-en-nlp}.
\begin{itemize}
\item\textbf{NLTK}
\newline
For a long time, NLTK (Natural Language ToolKit) has been the standard Python library for NLP. It combines algorithms for classifications, Part-of-Speech Tagging, stemming, tokenization (in) of words and sentences. It contains also data corpora and allows for a  sentiment analysis and  NER recognition .
\newline
\newline
Even if there are regular updates, the NLTK library starts a bit to date (2001) and shows some limits in particular in terms of performance.
\newline
\item\textbf{Spacy}
\newline
A newer library (2015) seems to have taken over from NLTK, it's about Spacy. This library written in Python and Cython includes the same types of tasks as NLTK: Tokenization, lemmatization, POS-tagging, feelings analysis (still in development), NER. 
It also has pre-trained word vectors and statistical models in several languages (English, German, French and Spanish so far).
\end{itemize}
\newpage
\subsubsection{Open information extraction}
\begin{figure}[ht]
  \begin{center}
  \mbox{\includegraphics[width=18cm, height=10cm]{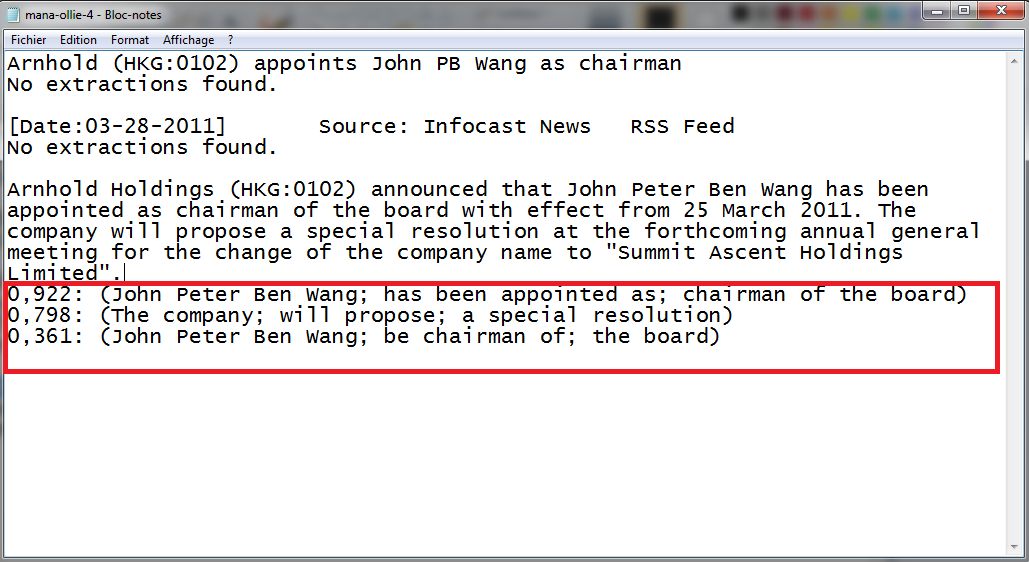}}
      \caption{Relationship triplet extraction}
    \label{Relationship triplet extraction}
    \end{center}
\end{figure}

We chose OLLIE as an OIE to extract relations\footnote{https://github.com/knowitall/ollie}.
To run OLLIE we should download  the latest OLLIE binary and the linear English MaltParser model (engmalt.linear-1.7.mco) and place it in the same directory as OLLIE.
We should run \textbf{java -Xmx512m -jar ollie-app-latest.jar yourfile.txt}.
Figure \ref{Relationship triplet extraction} clears the step of an open information extraction in order to extract a set of relationship triplets.

\subsubsection{Named entity recognition}
\begin{figure}[ht]
  \begin{center}
  \mbox{\includegraphics[width=18cm, height=9cm]{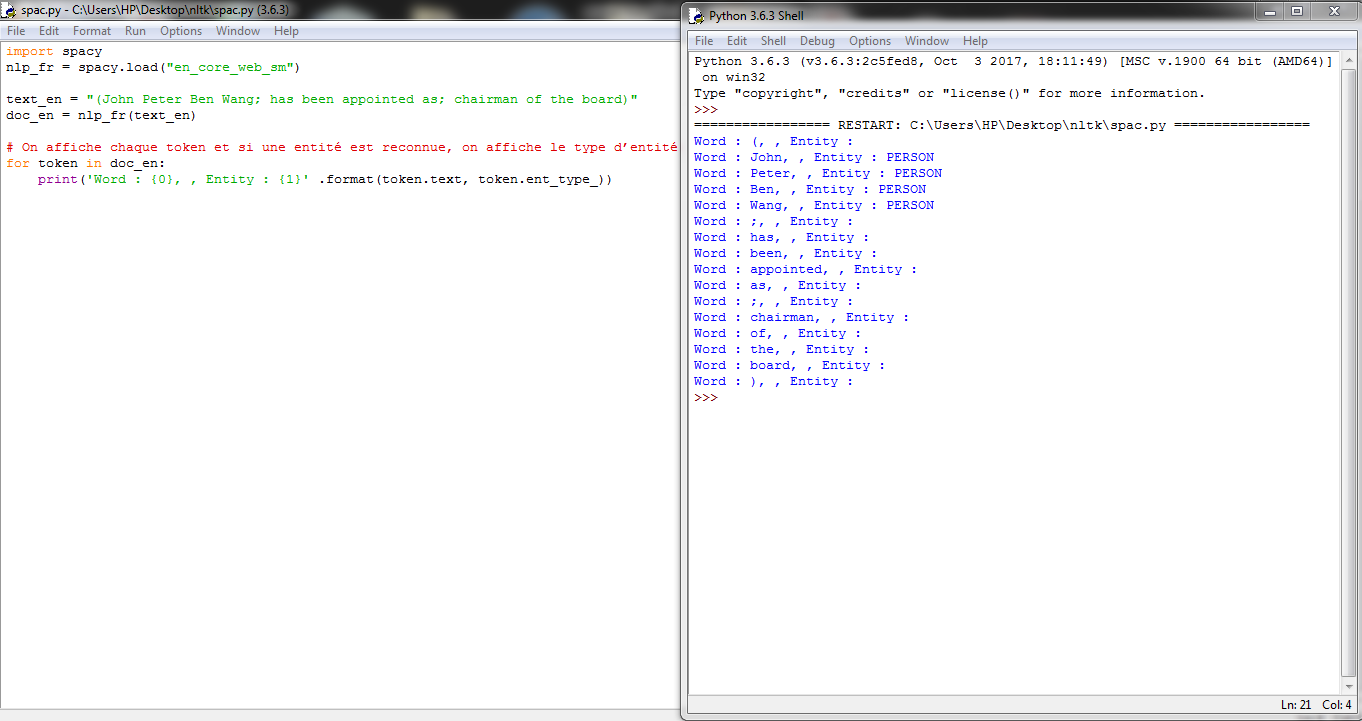}}
      \caption{NER with Spacy }
    \label{NER with Spacy}   
  \end{center}
\end{figure}

Figure \ref{Named entity recognition application} illustrates the recognition of named entities phase in our system applied to a text.
During  this phase we used the following two libraries: NLTK and Spacy \cite{maReference47}.
\newline
\begin{itemize}
\item There are 9 types of entities in NLTK: ORGANIZATION, PERSON, LOCATION, DATE, TIME, MONEY, PERCENT, FACILITY and GPE.
\newline
\item Spacy has 17 recognized entity types: PERSON, NORP, FACILITY, ORG, GPE, LOC, PRODUCT, EVENT, WORK\_OF\_ART, LANGUAGE, DATE, TIME, PERCENT, MONEY, QUANTITY, ORDINAL and CARDINAL. 
\end{itemize}
That's why we chose Spacy for NER and NLTK for lemmatisation.\newline
Figure \ref{NER with Spacy} shows the source code we used during this step.\newline
As we see the type POSITION is not among the recognized entities in Spacy that's why we have entered a list of positions in our system.
\newline
\newpage
\begin{figure}[ht]
  \begin{center}
  \mbox{\includegraphics[width=18cm, height=6cm]{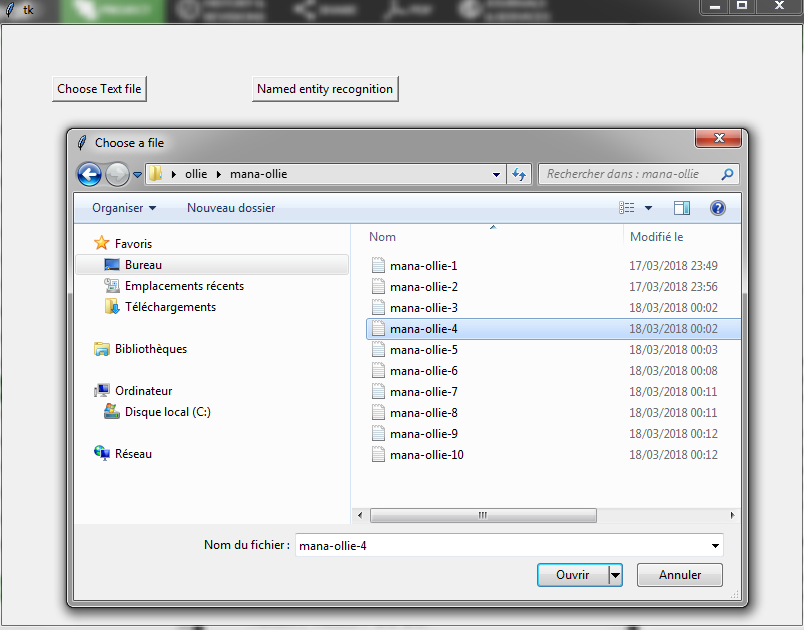}}
      \caption{Choose file application }
    \label{Choose file application}   
    \end{center}
\end{figure}
\begin{figure}[ht]
  \begin{center}
  \mbox{\includegraphics[width=18cm, height=6.5cm]{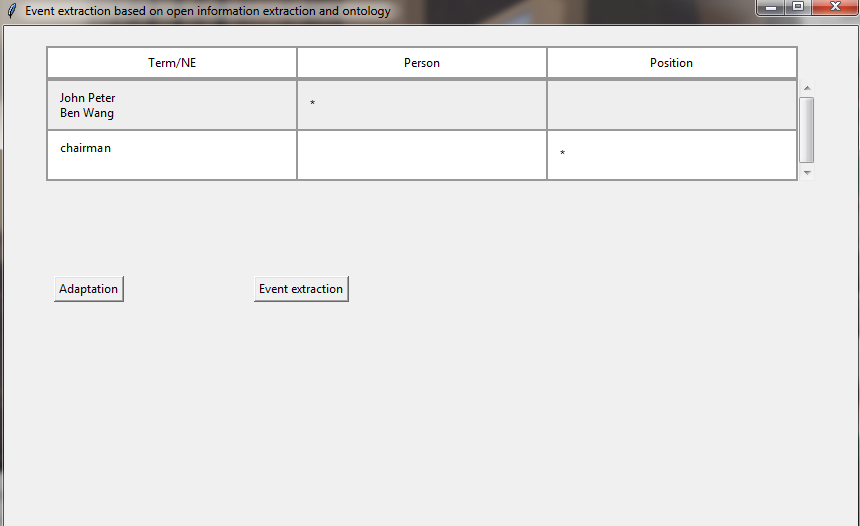}}
      \caption{Named entity recognition application }
    \label{Named entity recognition application}   
  \end{center}
\end{figure}
\newpage
\subsubsection{Ontology input adaptation}
\begin{figure}[ht]
 \begin{center}
\mbox{\includegraphics[width=18cm, height=9cm]{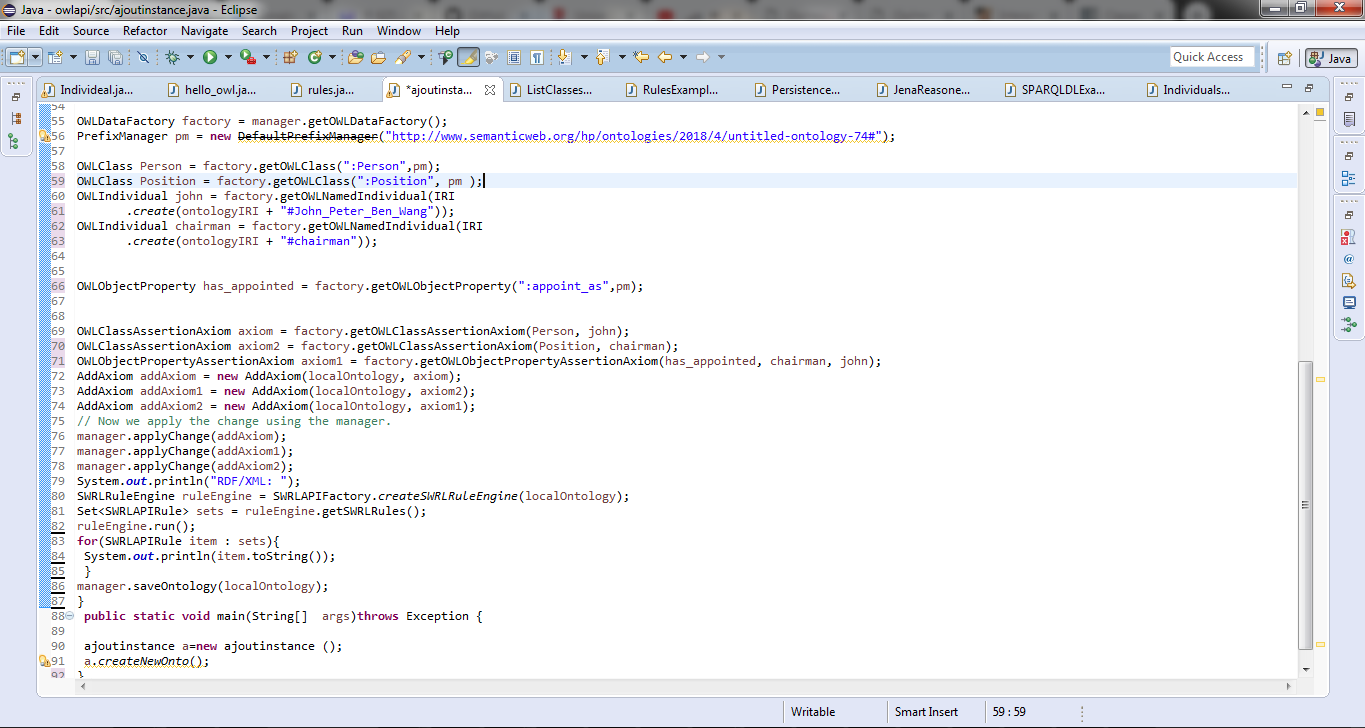}}
      \caption{Java code for adding instances to the ontology }
    \label{Java code for adding instances to the ontology}   
  \end{center}
\end{figure}

\begin{figure}[ht]
 \begin{center}
    \mbox{\includegraphics[width=18cm, height=6cm]{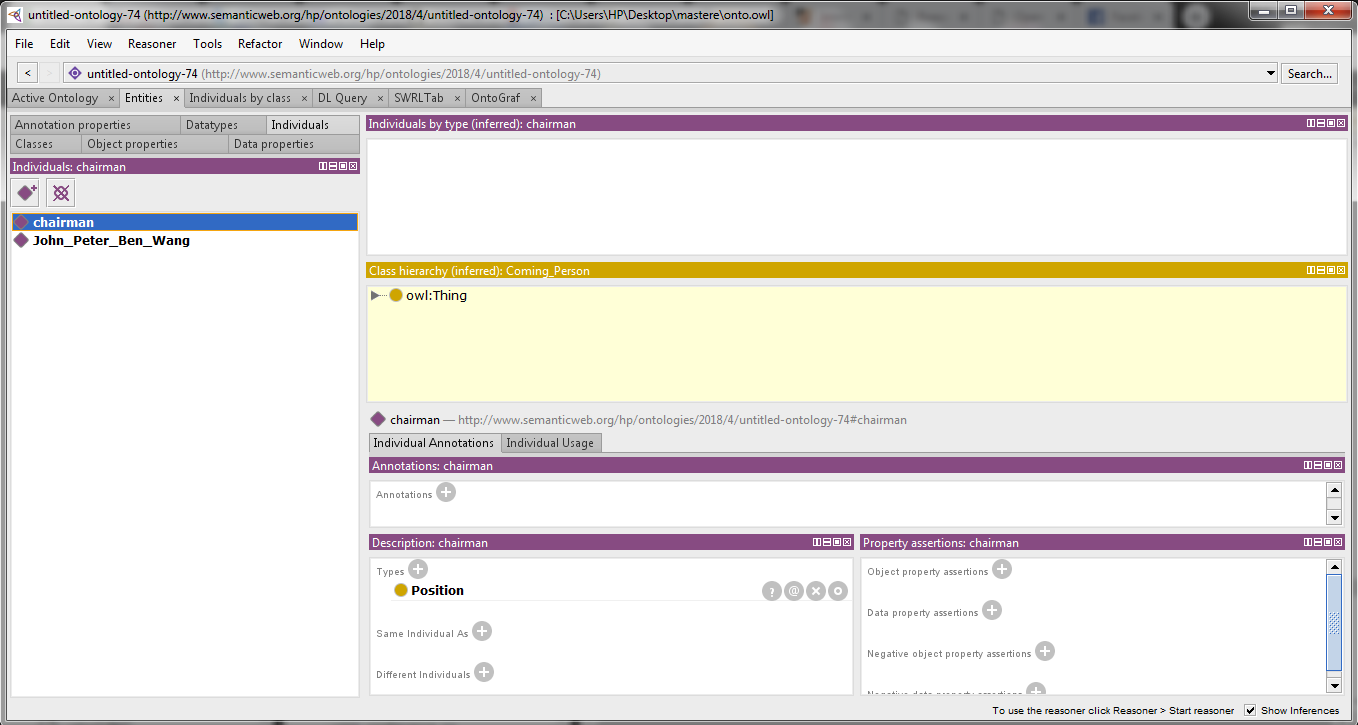}}
      \caption{Adaptation phase1 }
    \label{Adaptation phase1}   
  \end{center}
\end{figure}

To automatically add the instances (individuals) to the ontology  and to link them by their relationship, 
there are many APIs which provide objects and functions for manipulating
the elements that compose an ontology (i.e. classes, individuals,
properties, annotations, restrictions, etc). 
APIs exist, such as OWLAPI 
in Java, or other languages including Owlready2 in Python.\cite{maReference55}
We used a java code for adding the instances to the ontology as shown in Figure \ref{Java code for adding instances to the ontology} and python for the reasoning step.
We used Owlready2 as a module for ontology-oriented programming in Python. It can load OWL 2.0 ontologies as Python objects, modify them, save them, and perform reasoning via Hermit (included).
\newline
\newpage
Figure \ref{Adaptation phase1} shows what the ontology gives us during the adaptation phase.
\begin{figure}[ht]
  \begin{center}
    \mbox{\includegraphics[width=18cm, height=6cm]{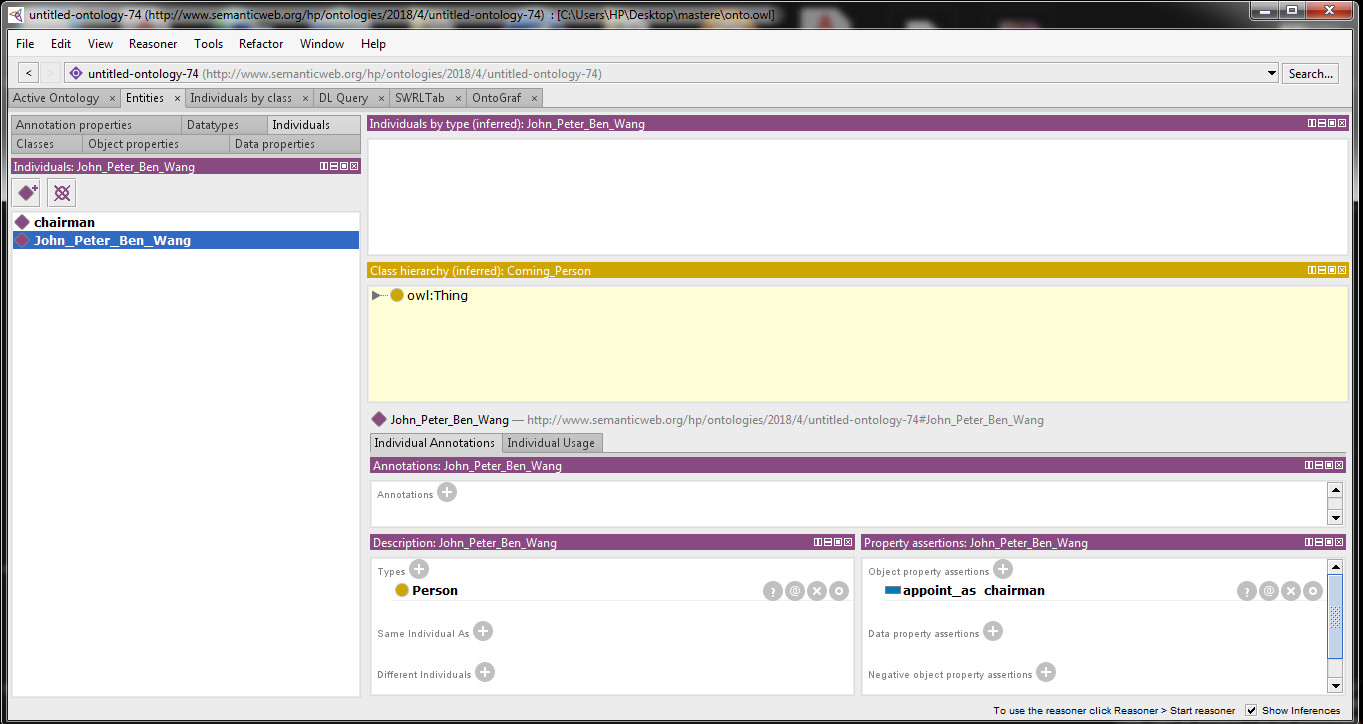}}
      \caption{Adaptation phase2 }
    \label{Adaptation phase2}   
  \end{center}
\end{figure}
\newpage
\begin{figure}[ht]
Figure \ref{Hermit reasoner } shows the Hermit reasoner included in protégé.
  \begin{center}
    \mbox{\includegraphics[width=18cm, height=10cm]{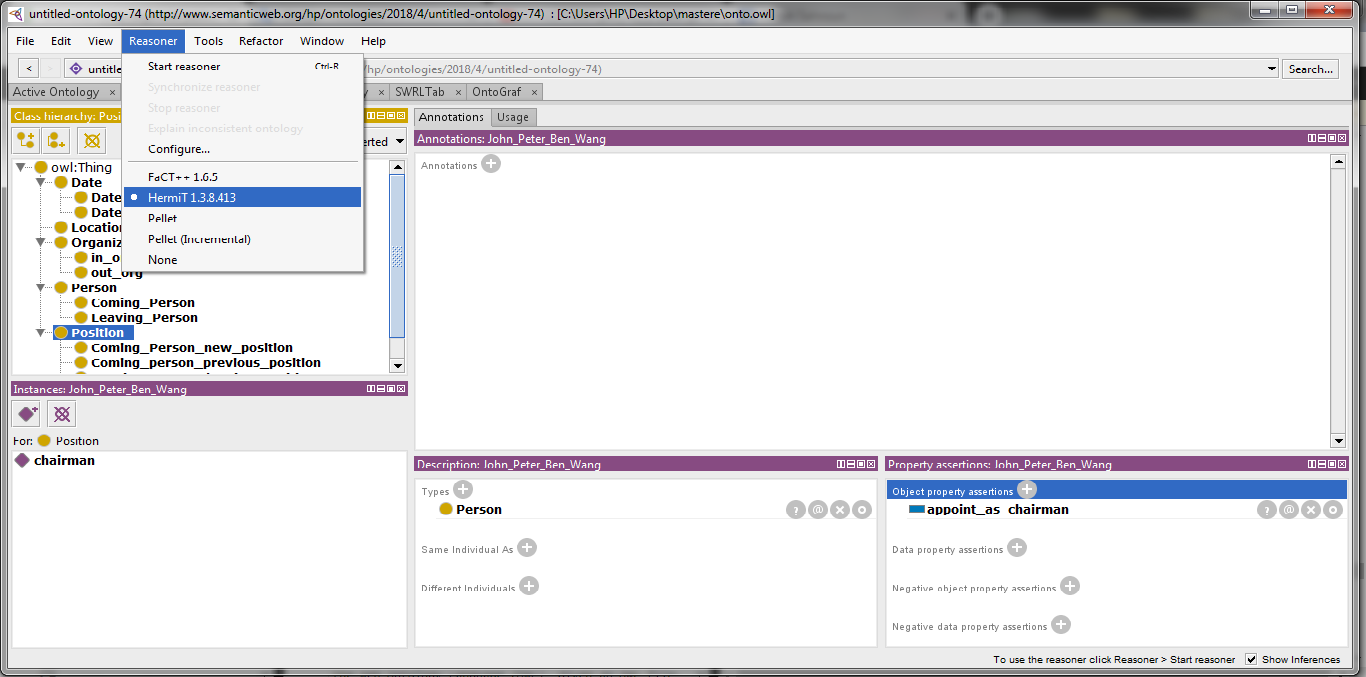}}
      \caption{Hermit reasoner }
    \label{Hermit reasoner }   
  \end{center}
\end{figure}
\subsubsection{Event extraction}
To extract  events we used the Hermit reasoner with the rules we created during the learning phase to affect for each input (individual) its role automatically.
Hermit is a reasoner for ontologies written using the Web Ontology Language (OWL). Given an OWL file, Hermit can determine whether or not the ontology is consistent, identify subsumption relationships between classes, and much more.
The reasoner is simply run by calling the sync\_reasoner() method of the ontology:
\begin{figure}[ht]
 \begin{center}
    \mbox{\includegraphics[width=4cm, height=1cm]{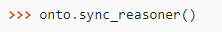}}
  \end{center}
\end{figure}
\newpage
Figure \ref{Running reasoner in protégé} shows how running hermit directly in protégé.
\begin{figure}[ht]
  \begin{center}    \mbox{\includegraphics[width=18cm, height=10cm]{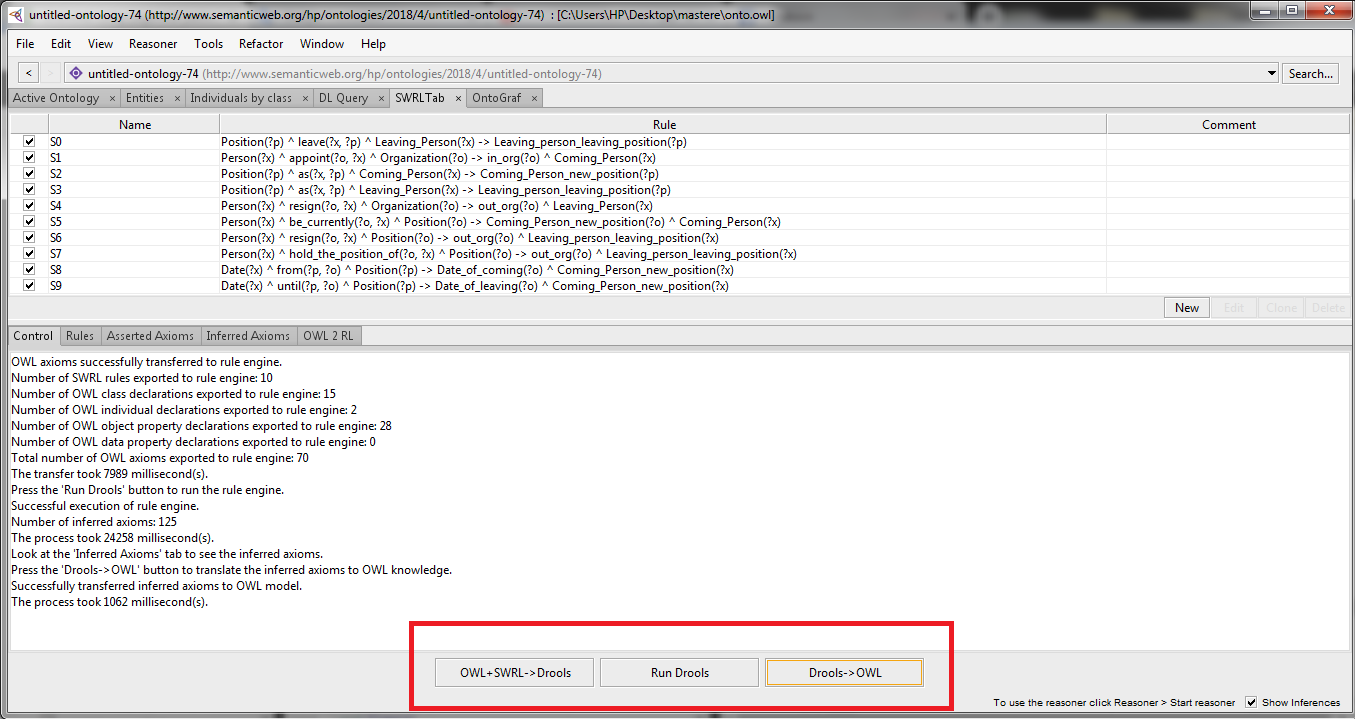}}
      \caption{Running reasoner in protégé }
    \label{Running reasoner in protégé}   
  \end{center}
\end{figure}
\begin{itemize}
\item Press the 'OWL+SWRL $\rightarrow$ Drools' button to transfer SWRL rules and relevant OWL knowledge to the rule engine.\newline
\item Press the 'Run Drools' button to run the rule engine.\newline
\item Press the 'Drools $\rightarrow$ OWL' button to transfer the inferred rule engine knowledge to OWL knowledge.
\end{itemize}
\newpage
\begin{figure}[ht]
Figure \ref{Assigning roles through the reasoner} shows the event extraction directly in protégé.
  \begin{center}
    \mbox{\includegraphics[width=18cm, height=6cm]{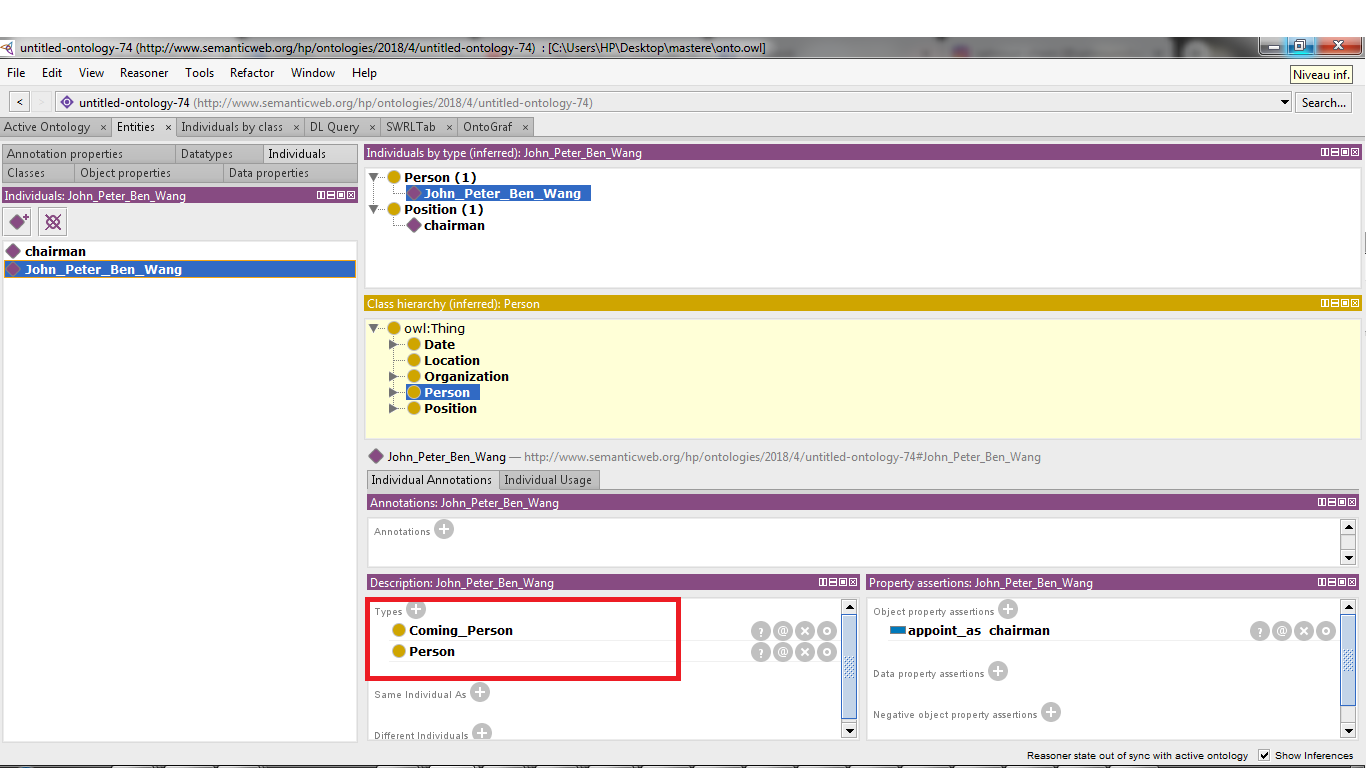}}
      \caption{Assigning roles through the reasoner }
    \label{Assigning roles through the reasoner}   
  \end{center}
  \end{figure}
\begin{figure}[ht]
  \begin{center}
    \mbox{\includegraphics[width=18cm, height=5cm]{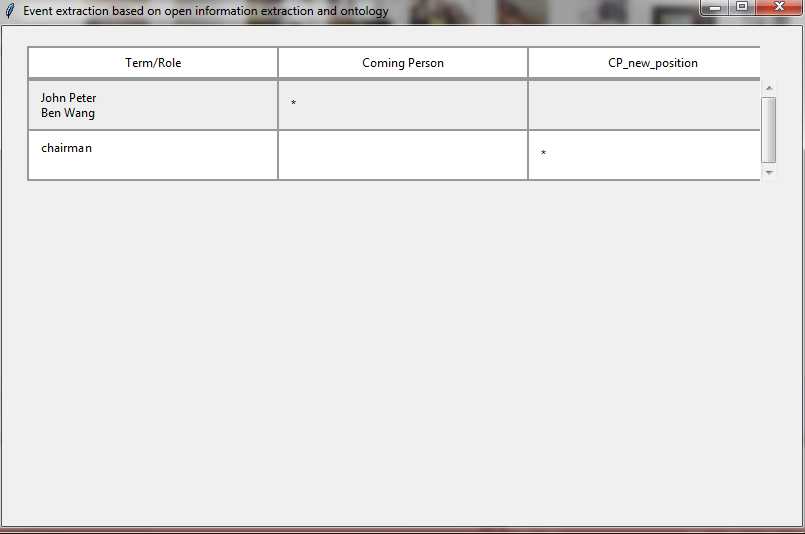}}
      \caption{Event extraction application }
    \label{Event extraction application}   
  \end{center}
\end{figure}
\newpage
\section{Evaluation}
We performed an evaluation to measure  the application's quality for the event recognition on a set of a "management change" corpora.
\subsection{Experimentation result}
The experimental study was carried out according to the three measurements of precision, recall and  F-mesure, which gave the following results in Table 4.1.
\newpage
\begin{table}[ht!]
\centering
\begin{tabular}{|M{1cm}|M{1.5cm}|M{1,5cm}|M{2cm}|}
\hline
\textbf{File}& \textbf{Precision}&\textbf{Recall}&\textbf{F-measure} \\
\hline
\textbf{F1}& {100\%}&{33\%}&{49\%} \\
\hline
\textbf{F2}& {100\%}&{85\%}&{91\%} \\
\hline\textbf{F3}& {100\%}&{88\%}&{93\%} \\
\hline\textbf{F4}& {100\%}&{50\%}&{66\%} \\
\hline\textbf{F5}& {100\%}&{66\%}&{80\%} \\
\hline\textbf{F6}& {77\%}&{66\%}&{71\%} \\
\hline\textbf{F7}& {94\%}&{94\%}&{94\%} \\
\hline\textbf{F8}& {100\%}&{100\%}&{100\%} \\
\hline\textbf{F9}& {100\%}&{85\%}&{91\%} \\
\hline
\textbf{F10}&{100\%}&{50\%}&{66\%} \\
\hline
\textbf{F11}&{100\%}&{77\%}&{82\%} \\
\hline\textbf{F12}& {100\%}&{75\%}&{85\%} \\
\hline\textbf{F13}& {100\%}&{57\%}&{72\%} \\
\hline\textbf{F14}& {100\%}&{80\%}&{88\%} \\
\hline\textbf{F15}& {100\%}&{60\%}&{75\%} \\
\hline\textbf{F16}& {0\%}&{0\%}&{0\%} \\
\hline\textbf{F17}& {100\%}&{75\%}&{85\%} \\
\hline\textbf{F18}& {75\%}&{25\%}&{37\%} \\
\hline
\textbf{F19}&{0\%}&{0\%}&{0\%} \\
\hline
\textbf{F20}&{88\%}&{100\%}&{93\%} \\
\hline\textbf{F21}& {83\%}&{30\%}&{44\%} \\
\hline\textbf{F22}&{0\%}&{0\%}&{0\%} \\
\hline\textbf{F23}&{100\%}&{100\%}&{100\%} \\
\hline\textbf{F24}& {100\%}&{28\%}&{43\%} \\
\hline\textbf{F25}&{60\%}&{60\%}&{60\%} \\
\hline\textbf{Total}&\textbf{85\%}&\textbf{59\%}&\textbf{66\%} \\
\hline
\end{tabular}
\caption{Experimental study}
\label{table:6}
\end{table}
The results range from 0\% to 100\% due to the accuracy of the OIE system, i.e sometimes OIE can detect the triplet which has the  targeted information and sometimes no. 
\newline
\newline
To establish a comparative study with an other work in the field of event extraction, the two-level event extraction approach is based on CRFs, so we chose to compare our results against this one.
Precision is considered to be a comparative metric between the two approaches.
Table 4.2 presents the results obtained from these two approaches.
\begin{table}[ht!]
\centering
\begin{tabular}{|M{4cm}|M{4cm}|M{5cm}|}
\hline
\textbf{Role}&\textbf{Our approach}&\textbf{Approach of two levels of event extraction} \\
\hline
\textbf{Coming\_Person} & 81\% &81\% \\
\hline
\textbf{Leaving\_Person} & 92\%   & 49\%\\
\hline
\textbf{Date\_of\_coming} & 22\% &69\%  \\
\hline
\textbf{Date\_of\_leaving} & 50\% &61\% \\
\hline
\textbf{CP\_new\_position} & 60\% &81\%  \\
\hline
\textbf{LP\_leaving\_position} & 75\% &60\%  \\
\hline
\end{tabular}
\caption{Comparative study
}
\label{table:7}
\end{table}
\subsection{Discussion}
According to the tables, the results we have obtained are quite satisfactory and encouraging. However we can deduce that we arrived with an acceptable rate to extract from the texts a set of events. The two-level event extraction approach based on CRF analysis that requires human and manual intervention at the classifiers generation  and which take a long time while our approach everything is was automatic except that part of the construction of the rules.
\section{Conclusion}
In this chapter, we have focused our interest on the test results. In this context we presented our application and we described more particularly the different steps and the tools used in each one, then we evaluated the results of the test obtained and compare them by another system which is based on two levels event extraction approach using CRFs.
\chapter*{Conclusion and perspectives}
\markboth{Conclusion and perspectives}{}
\addcontentsline{toc}{chapter}{Conclusion and perspectives}
Access to useful information from a large amount of data has given rise to several fields of IE. Our study is a part of an emerging field of IE: It is the event extraction.
\newline \newline
In this context, we have created an event extraction system from textual documents. We followed a learning phase process using an ontology and a recognition phase using OIE, NER, and an adaptation step through a reasoner.
\newline \newline
The main objective of this work is to construct a system for an automatic extraction of management change event from texts.  we compared our work to a previous application that is based on a correspondence between NEs and  events. This correspondence is based on a double generation of the classifier: A classifier for the first level's learning (PERSON, POSITION, ORG) and another classifier for the second level's learning (COMING PERSON, NEW POSITION, IN\_ORG)
which has had good results with a 53\% of recall and a 71\% of precision.
\newline 
\newline
In addition a second application has been applied based on our approach.  The results we have obtained are encouraging: 85\% of precision and 59\% of recall. 
\newline
\newline
This result is achieved thanks to the different aspects of this work:
First of all, we clarified some basic notions related to our field of research. We also reviewed the theoretical framework analysis in  the field of OIE and  IE,  and more particularly the  event extraction.
\newline
This analysis has given us a much clearer vision of our problem.
Among other things, this allowed us to position our work with respect to the state of the art.
\newline
\newline
To implement our event extraction system, we had the choice between linguistic, statistical and hybrid techniques. Regarding to the recognition part, our research work proposes
the adaptation  which we merge between NEs and  event through the ontology, the rules construction and the reasoner.
\newline
\newline
Based on the evaluation of our event extraction system, we believe that the results obtained are encouraging compared to the results  of two-level event extraction approach.
\newline
\newline
The work done in this master thesis orient us towards two axes of perspectives:
\begin{itemize}
\item We can merge between OIE and open domain event extraction approach\cite{maReference46}. Based on the idea of the open domain we can realize another approach on this context and it can give us more effective results.
\newline
\newline
\item We have worked with lemmatized verbs but we can also work  on conjugated verbs by giving importance to the attribution / condition part for the second type of OIE system to resolve the problem of two different roles in the same sentence and to identify the temporal order of statements.
\newline
\newline
\item We can merge between triplets which has got common tokens to extract as much information as possible.

\nocite{ref1}\nocite{ref2}\nocite{ref3}\nocite{ref4}\nocite{ref5}\nocite{ref6}\nocite{ref7}\nocite{ref8}\nocite{ref9}\nocite{ref10}\nocite{ref11}\nocite{ref12}\nocite{ref13}\nocite{ref14}\nocite{ref15}\nocite{ref16}\nocite{ref17}\nocite{ref18}\nocite{ref19}\nocite{ref20}

\end{itemize}
\renewcommand{\bibname}{Bibliography}

\end{spacing}
\newpage
\thispagestyle{empty}
\textRL{\textbf{
استخراج الحدث على أساس استخراج المعلومات المفتوحة والأنطولوجيا\qquad\qquad\quad} }
\newline \textRL{\textbf{
الملخص.} يتضمن  هذا العمل المقدم في أطروحة الماجستير  استخراج مجموعة من الأحداث من النصوص المكتوبة باللغة الطبيعية. لهذا الغرض ، اعتمدنا على المفاهيم الأساسية لاستخراج المعلومات وكذلك استخراج المعلومات المفتوحة. أولاً ، طبقنا نظامًا  لاستخراج المعلومات المفتوحة  لهدف استخراج العلاقات ، لإبراز أهمية استخراج المعلومات المفتوحة في استخراج الحدث ، واستخدمنا الأنطولوجيا في نمذجة الحدث. اختبرنا النتائج  باستخدام مقاييس الاختبار. ونتيجة لذلك ، أظهر منهج استخراج الحدث المكون من مستويين نتائج أداء جيدة ولكنه يتطلب الكثير من تدخل الخبراء في بناء المصنفات وهذا سيستغرق بعض الوقت. ، في هذا السياق ، اقترحنا مقاربة تقلل من تدخل الخبراء في علاقة الاستخراج الذي يكون آليًا  ، لإثبات صلة النتائج المستخلصة ، أجرينا سلسلة من التجارب باستخدام مقاييس اختبار     مختلفة بالإضافة إلى دراسة مقارنة.\qquad\qquad\qquad\quad\qquad\qquad\qquad\quad\qquad\qquad\qquad\quad\qquad\qquad\qquad\quad\qquad\qquad\qquad\quad 
\textbf{الكلمات المفاتيح.}  \quadاستخراج المعلومات ،  الحدث  ،  الكيان المسمى ، العلاقة ، استخراج المعلومات المفتوحة ، أونطولوجيا.
}
\newline
\begin{center}\textbf{ 
Extraction d'événements basée sur l'extraction d'informations ouverte et l'ontologie 
}\end{center}
\textbf{Résumé.} Le travail présenté dans ce mémoire consiste à extraire un ensemble d'évènements à partir des textes écrits en langage naturel. Pour cet objectif, nous avons adopté les notions de base d'extraction d'information ainsi que l'extraction d'information ouverte. Tout d'abord, nous avons appliqué un système d'extraction d'information ouverte(EIO) pour l'extraction des triplets de relations , pour mettre en valeur l'importance des EIO dans l'extraction d'évènement et on a utilisé l'ontologie pour la modélisation de l'évènement. Nous avons testé les résultats de notre approche avec des métriques de tests. En conséquence, l'approche d'extraction d'évènement à deux niveaux a montré des bonnes résultats de performances mais nécessite l'itervention de l'expert dans la construction des classifieurs et ça va prendre du temps. Dans ce cadre nous avons proposé une approche qui diminue l'intervention de l'expert au niveau d'extraction des relations, la reconnaissance des entités nommées et au raisonnement en se basant sur des techniques d'adaptation et de correspondance. Enfin pour prouver la pertinence des résultats extraits, nous avons mené une série d'expérimentations en utilisant différentes métriques de tests  ainsi qu'une étude comparative.\newline
\textbf{Mots clés.} Extraction d'informations,  Evènement, Entité nommée, Relation, OIE, Ontologie
\newline
\begin{center}
\textbf{ \quad\quad\quad Event extraction based on open information extraction and ontology}
\end{center}
\textbf{Abstract.} The work presented in this master thesis consists of extracting a set of events from texts written in natural language. For this purpose, we have based ourselves on the basic notions of the information extraction as well as the open information extraction. First, we applied an open information extraction(OIE) system for the relationship  extraction, to highlight the importance of OIEs in event extraction, and we used the ontology to the  event modeling. We tested the results of our approach with test metrics. As a result, the two-level event extraction approach has shown good performance results but requires a lot of expert intervention in the construction of classifiers and this will take time. In this context we have proposed an approach that reduces  the expert intervention  in the relation extraction, the recognition of entities and the reasoning which are automatic and based on techniques of adaptation and correspondence. Finally, to prove the relevance of the extracted results, we conducted a set of experiments using different test metrics as well as a comparative study.
\newline
\textbf{Keywords.} Information extraction, Event, Named Entity, Relationship, OIE, Ontology.

\begin{thebibliography}{9}
\addcontentsline{toc}{chapter}{Bibliography}
\bibitem{maReference1}
Grishman R and Sundheim B. Message understanding conference-6: a brief history. \emph{In: Proceedings of the 16th conference on Computational linguistics}, 466-471, (1996)
\bibitem{maReference2}
Meryem Talha, Siham Boulaknadel,and Driss Aboutajdine. \emph{Système de Reconnaissance des Entités Nommées Amazighes},Conference: Traitement Automatique des Langues Naturelles, (2014).
\bibitem{maReference3}
Samir Elloumi, Ali Jaoua, Fethi Ferjani, Nasredine Semmar, Romaric Besançon, Jihad Al-Jaam and Helmi Hmmami. \emph{General learning approach for event extraction: Case of management change event} Research Article,page(s): 211-224,(2012) 
\bibitem{maReference4}
Yayoi Nakamura-Delloye, Rosa Stern. \emph{Extraction de relations et de patrons de relations entre entités nommées en vue de l’enrichissement d’une ontologie},Conference papers  2011 : Terminologie \& Ontologie : Théories et Applications, May 2011, Annecy, France. pp.50 (2011).
\bibitem{maReference5}
B.V. Durme, L.K. Schubert,  \emph{Open knowledge extraction using compositional language processing}. In Proceeding
STEP '08 Proceedings of the 2008 Conference on Semantics in Text Processing, Pages 239-254  (2008).
\bibitem{maReference6}
O. Etzioni, M. Banko, S. Soderland, D.S. Weld, \emph{Open Information Extraction from the Web. Communications of the ACM, December 2008, Vol. 51 No. 12, Pages 68-74.}
(2008).
\bibitem{maReference7}
Jiang J and Zhai C. \emph{A systematic exploration of the feature space for relation extraction.},Human language technologies 2007: The conference of the North American Chapter of the Association for Computational Linguistics; 22 - 27 April 2007, Rochester, New York, 113-120, (2007).
\bibitem{maReference8}
Frederik Hogenboom, Flavius Frasincar, Uzay Kaymak, and Franciska de Jong. \emph{An Overview of Event Extraction from Text},Proceedings of Detection, Representation, and Exploitation of Events in the Semantic Web (DeRiVE 2011), Workshop in conjunction with the 10th International Semantic Web Conference 2011 (ISWC 2011), Bonn, Germany, October 23,Pages 48-57, 2011.
\bibitem{maReference9}
Florence Amardeilh.  \emph{Web sémantique et Informatique Linguistique: propositions méthodologiques et réalisation d'une plateforme logicielle.}, Thèse domain\_stic.gest. Université de Nanterre - Paris X, 2007. Français  (2007).
\bibitem{maReference10}
D. J. Allerton. \emph{Stretched Verb Constructions in English.
Routledge Studies in Germanic Linguistics},318 pages, (2002).
\bibitem{maReference11}
A. Fader, S. Soderland, O. Etzioni, \emph{Identifying Relations for
Open Information Extraction}, Proceedings of the Conference on Empirical Methods in Natural Language Processing.Pages 1535-1545 (2011).
\bibitem{maReference12}
L.D. Corro, R. Gemulla. ClausIE: Clause-Based Open
Information Extraction. WWW '13 Proceedings of the 22nd international conference on World Wide Web.
Pages 355-366.(2013)
\bibitem{maReference13}
M. Banko, M.J. Cafarella, S. Soderland, M. Broadhead, O.
Etzioni, O, \emph{Open Information Extraction from the Web.} In
IJCAI'07 Proceedings of the 20th international joint conference on Artifical intelligence.Pages 2670-2676, (2007).
\bibitem{maReference14}
F. Wu, D.S Weld, \emph{Open Information Extraction using
Wikipedia.} In Proceedings Proceeding
ACL '10 Proceedings of the 48th Annual Meeting of the Association for Computational Linguistics.Pages 118-127  (2010).
\bibitem{maReference15}
Duc-Thuan Vo and Ebrahim Bagheri.\emph{ Open Information Extraction}, (2016).
\bibitem{maReference16}
L.D. Corro, R. Gemulla, \emph{ ClausIE: Clause-Based Open
Information Extraction.} In Proceedings of the 22nd international conference on World Wide Web, page 355-366, (2013).
\bibitem{maReference17}
R. Quirk, S. Greenbaum, G. Leech, J. Svartvik,\emph{ A
Comprehensive Grammar of the English Language.} Research Article ,page(s): 122-136, (1985).
\bibitem{maReference18}
Sowa, J.F .\emph{The Challenge of Knowledge Soup. In Proc. First International WordNet Conference. 55-90} (2004).
\bibitem{maReference19}
Ferrucci, D.A .\emph{ IBM's Watson / DeepQA. In: Proceedings of the 38th Annual International Symposium on Computer Architecture.} ACM , Pages 365-376, (2011).
\bibitem{maReference20}
Andrew Hsi.\emph{ Event Extraction for Document-Level Structured Summarization}PHD thesis,( 2017). 
\bibitem{maReference21}
Sachin Pawar, Girish K. Palshikar, Pushpak Bhattacharyya.\emph{"Relation Extraction : A Survey}.(2017).
\bibitem{maReference22}
Daniel Jurafsky , James H. Martin.\emph{Speech and Language Processing}.Pages 36-51,(2017)
\bibitem{maReference23}
Volcani, Yanon . Fogel, David B.\emph{System and method for determining and controlling the impact of text}.(2001).
\bibitem{maReference24}
Pauline Olivier. \emph{Introduction to NLP}(2017).
\bibitem{maReference25}
Pauline Olivier .\emph{Simple NLP tasks tutorial
}(2017).
\bibitem{maReference26}
Pauline Olivier .\emph{Le nouvel essor du Natural Language Processing – NLP
}(2018).
\bibitem{maReference27}
S Kahane.\emph{WHY TO CHOOSE DEPENDENCY
RATHER THAN CONSTITUENCY FOR SYNTAX:A FORMAL POINT OF VIEW }, 257-272, (2012)
\bibitem{maReference28}
Satoshi Sekine,
Kiyoshi Sudo,
Chikashi Noba.\emph{Extended Named Entity Hierarchy },Pages 1818- 1824, (2002)
\bibitem{maReference29}
SHAALAN  K.,  RAZA  H.    \emph{NERA  :  Named  entity  recognition  for  arabic.  Journal  OF  the  American  Society  for Information Science and Technology},Pages 1652–1663,(2009)
\bibitem{maReference30}
ZAGHOUANI  W.,  POULIQUEN  B.,  EBRAHIM  M.,  STEINBERGER  R.   \emph{Adapting  a  resource-light  highly multilingual named entity recognition system to arabic.} Proceedings of the Seventh conference on International Language Resources and Evaluation, Pages 563-567,(2010)
\bibitem{maReference31}
BENAJIBA Y, DIAB M, ROSSO P. \emph{Using Language Independent and Language Specific Features to
Enhance Arabic Named Entity Recognition.} The International Arab Journal of Information Technology,Pages 464-473
(2009)
\bibitem{maReference32}
ABULEIL S. \emph{Hybrid System for Extracting and Classifying Arabic Proper Names.} Conf. on Artificial Intelligence, Knowledge Engineering and Data Bases, Madrid-Spain, Pages 205-210 ,(2006)
\bibitem{maReference33}
P Srikanth , Kavi Narayana Murthy. \emph{Named Entity Recognition for Telugu.},In Proceedings of the IJCNLP-08 Workshop on NER for South and South East Asian Languages,Pages 41-50, (2008)
\bibitem{maReference34}
E. Agichtein and L. Gravano. \emph{Snowball: Extracting Relations
from Large Plain-Text Collections.}, Proceedings of the fifth ACM conference on Digital libraries, Pages 85-94, (2000).
\bibitem{maReference35}
Marti A. Hearst. \emph{Automatic acquisition of hyponyms
from large text corpora.}, Proceeding
COLING '92 Proceedings of the 14th conference on Computational linguistics ,Pages 539-545,  (1992)
\bibitem{maReference36}
Matthew Berland, Eugene Charniak .\emph{Finding Parts in Very Large Corpora.},Proceedings of the 37th annual meeting of the Association for Computational Linguistics on Computational Linguistics
Pages 57-64,(1999)
\bibitem{maReference37}
Brin,S.\emph{Extracting Patterns and Relations from the World Wide Web}(1998)
\bibitem{maReference38}
Eugene Agichtein,Luis Gravano \emph{Snowball: extracting relations from large plain-text collections}(2000)
\bibitem{maReference39}
Sunita Sarawagi and William W. Cohen.\emph{Semi-Markov
Conditional Random Fields for Information
Extraction.},226-241,(2004).
\bibitem{maReference40}
Niu,F. et al. \emph{Elementary: large-scale knowledge-base construction
via machine learning and statistical.}, International Journal on Semantic Web and Information Systems (IJSWIS). Pages 1-23, (2012)
\bibitem{maReference41}
Gabor Angeli, Julie Tibshirani, Jean Wu, and Christo-pher D Manning. \emph{Combining distant and partial supervision for relation extraction.},Pages 1556-1567, (2014)
\bibitem{maReference42}
Dorian Kodelja, Romaric Besancon, Olivier Ferret. \emph{Représentations et modèles en extraction d’événements
supervisée}(2017)
\bibitem{maReference43}
Samir Elloumi. \emph{An adaptive model for Event extraction.
A comparison between the CRF-based and GLA2E classiers},page(s): 211-224,(2016)
\bibitem{maReference44}
Sunita Sarawagi .William W. Cohen\emph{Semi-Markov Conditional Random Fields for Information Extraction}(2004)
\bibitem{maReference45}
Laurie Serrano, Maroua Bouzid, Thierry Charnois, Bruno Grilhères\emph{Vers un système de capitalisation des connaissances :
extraction d’événements par combinaison de plusieurs approches.}(2012)
\bibitem{maReference46}
Kang Liu. \emph{
Open Domain Event Extraction from Texts}.(2017)
\bibitem{maReference47}
Pauline Olivier. \emph{Simple NLP tasks tutorial}.(2017)
\bibitem{maReference48}
Tomas Mikolov, Ilya Sutskever, Kai Chen, Greg S Corrado,
and Jeff Dean. \emph{Distributed representations
of words and phrases and their compositionality.
In Advances in neural information processing
systems},1-9,(2013)
\bibitem{maReference49}
Nguyen, T. H., \& Grishman, R. \emph{Relation Extraction: Perspective from Convolutional Neural Networks. Workshop on Vector Modeling for NLP},39-48,(2015).
\bibitem{maReference50}
Jason P.C. Chiu, Eric Nichols\emph{Named Entity Recognition with Bidirectional LSTM-CNNs}(2016)
\bibitem{maReference51}
 Yubo Chen, Liheng Xu, Kang Liu, Daojian Zeng, Jun Zhao, et al.\emph{Event
Extraction via Dynamic Multi-Pooling Convolutional Neural Networks.}Pages 167-176,(2015) 
\bibitem{maReference52}
Shulin Liu, Yubo Chen, Kang Liu, and Jun Zhao.\emph{Exploiting Argument
Information to Improve Event Detection via Supervised Attention Mechanisms.
In Proceedings of the 55th Annual Meeting of the Association for Computational
Linguistics },P17-1164 (2017).
\bibitem{maReference53}
Thien Huu Nguyen, Kyunghyun Cho, and Ralph Grishman. \emph{Joint Event
Extraction via Recurrent Neural Networks},Pages 300-309, (2016).
\bibitem{maReference54}
Dorian Kodelja, Romaric Besancon, Olivier Ferret. \emph{Représentations et modèles en extraction d’événements supervisée}, (2017).
\bibitem{maReference55}
Jean-Baptiste Lamy. \emph{Owlready: Ontology-oriented programming in Python
with automatic classification and high level constructs
for biomedical ontologies},Artificial Intelligence in Medicine, Elsevier, pp.11 - 28. (2017).







\bibitem{ref1}Hamrouni, T., Ben Yahia, S. and Nguifo, E.M. Generalization of association rules through disjunction. \textit{Annals of Mathematics and Artificial Intelligence}, 59(2), pp.201-222, (2010).

\bibitem{ref2}Cellier, P., Ferré, S., Ridoux, O. and Ducasse, M. A parameterized algorithm to explore formal contexts with a taxonomy. \textit{International Journal of Foundations of Computer Science}, 19(02), pp.319-343, (2008).

\bibitem{ref3}Ben Yahia, S., and E. Mephu Nguifo. Revisiting generic bases of association rules. In \textit{International Conference on Data Warehousing and Knowledge Discovery}, pp. 58-67. Springer, Berlin, Heidelberg, (2004).

\bibitem{ref4}Koutsoni, Olga, Mourad Barhoumi, Ikram Guizani, and Eleni Dotsika. Leishmania eukaryotic initiation factor (LeIF) inhibits parasite growth in murine macrophages. \textit{PLoS One} 9, no. 5 (2014): e97319.

\bibitem{ref5}Djeddi, Warith Eddine, Mohamed Tarek Khadir, and Sadok Ben Yahia. XMap: results for OAEI 2015. In \textit{OM}, pp. 216-221. (2015).


\bibitem{ref6}Brahmi, Hanen, Imen Brahmi, and Sadok Ben Yahia. OMC-IDS: at the cross-roads of OLAP mining and intrusion detection. In \textit{Pacific-Asia Conference on Knowledge Discovery and Data Mining}, pp. 13-24. Springer, Berlin, Heidelberg, (2012).

\bibitem{ref7}Mouakher, Amira, and Sadok Ben Yahia. Qualitycover: efficient binary relation coverage guided by induced knowledge quality. \textit{Information Sciences} 355 (2016): 58-73.

\bibitem{ref8}Hamdi, Sana, Alda Lopes Gancarski, Amel Bouzeghoub, and Sadok Ben Yahia. IRIS: A novel method of direct trust computation for generating trusted social networks. In \textit{2012 IEEE 11th International Conference on Trust, Security and Privacy in Computing and Communications}, pp. 616-623. IEEE, (2012).

\bibitem{ref9}Djeddi, Warith Eddine, Mohamed Tarek Khadir, and S. Ben Yahia. XMap++: results for OAEI 2014. In \textit{OM}, pp. 163-169. (2014).

\bibitem{ref10}Jelassi, M. Nidhal, Christine Largeron, and Sadok Ben Yahia. Efficient unveiling of multi-members in a social network. \textit{Journal of Systems and Software} 94 (2014): 30-38.

\bibitem{ref11}Hamrouni, Tarek, S. Ben Yahia, and E. Mephu Nguifo. Looking for a structural characterization of the sparseness measure of (frequent closed) itemset contexts. \textit{Information Sciences} 222 (2013): 343-361.

\bibitem{ref12}Ben Yahia, Sadok, and Engelbert Mephu Nguifo. Emulating a cooperative behavior in a generic association rule visualization tool. In \textit{16th IEEE International Conference on Tools with Artificial Intelligence}, pp. 148-155. IEEE, (2004).

\bibitem{ref13}Sassi, Imen Ben, Sehl Mellouli, and Sadok Ben Yahia. Context-aware recommender systems in mobile environment: On the road of future research. \textit{Information Systems} 72 (2017): 27-61.

\bibitem{ref14}Zghal, Sami, Sadok Ben Yahia, Engelbert Mephu Nguifo, and Yahya Slimani. SODA: an OWL-DL based ontology matching system. In \textit{Proceedings of the 2nd International Conference on Ontology Matching}-Volume 304, pp. 261-267. CEUR-WS. org, (2007).

\bibitem{ref15}Hamrouni, Tarek, Sadok Ben Yahia, and Engelbert Mephu Nguifo. Succinct system of minimal generators: A thorough study, limitations and new definitions. In \textit{International Conference on Concept Lattices and Their Applications}, pp. 80-95. Springer, Berlin, Heidelberg, (2006).

\bibitem{ref16}Ben Yahia, S., and A. Jaoua. A top-down approach for mining fuzzy association rules. In \textit{Proc. 8th Int. Conf. Information Processing Management of Uncertainty Knowledge-Based Systems}, pp. 952-959. (2000).

\bibitem{ref17}Kachroudi, Marouen, Essia Ben Moussa, Sami Zghal, and Sadok Ben Yahia. Ldoa results for oaei 2011. In \textit{Proceedings of the 6th International Conference on Ontology Matching}-Volume 814, pp. 148-155. CEUR-WS. org, (2011).

\bibitem{ref18}Younes, Nassima Ben, Tarek Hamrouni, and Sadok Ben Yahia. Bridging conjunctive and disjunctive search spaces for mining a new concise and exact representation of correlated patterns. In \textit{International Conference on Discovery Science}, pp. 189-204. Springer, Berlin, Heidelberg, (2010).

\bibitem{ref19}Samet, Ahmed, Eric Lefevre, and Sadok Ben Yahia. Mining frequent itemsets in evidential database. In \textit{Knowledge and Systems Engineering}, pp. 377-388. Springer, Cham, (2014).

\bibitem{ref20}Gasmi, G., S. Ben Yahia, E. Mephu Nguifo, and Y. Slimani. IGB: une nouvelle base générique informative des regles d’association. \textit{Revue I3 (Information-Interaction-Intelligence)} 6, no. 1 (2006): 31-67.


\end{thebibliography}
\end{document}